%% file: iclr2027_conference.tex
\documentclass{article} 
\usepackage{iclr2027_conference,times}
\iclrfinalcopy
\input{math_commands.tex}

\usepackage{url}
\usepackage{graphicx}
\usepackage{float}
\usepackage{booktabs}
\usepackage{makecell}
\usepackage{todonotes}
\usepackage{hyperref} 
\usepackage{enumitem}

\title{Neural Operator Splitting for Fluid Dynamics in Complex Geometry }

\author{
Jianxiang Li\thanks{Equal contribution.} \\
Peking University \\
\texttt{lijianxiang0619@stu.pku.edu.cn}
\And
Xi Deng\footnotemark[1] \\
California Institute of Technology \\
\texttt{xideng94@caltech.edu}
\AND
Hong Chul Nam\footnotemark[1] \\
New York University \\
\texttt{hn2557@nyu.edu}
\And
Anima Anandkumar \\
California Institute of Technology \\
\texttt{anima@caltech.edu}
}

\usepackage{booktabs}
\begin{document}
\maketitle
\lhead{}

\maketitle

\begin{abstract}
Solving dynamical systems in complex and diverse geometries remains challenging for both numerical solvers and learning-based surrogate models. We propose Neural Operator Splitting, a framework for learning fluid dynamics while preserving the physical structure of fractional-step numerical schemes. Instead of learning the full-step evolution monolithically, we decompose each time step into physically interpretable steps, and learn each operator separately using intermediate states from the underlying numerical solver. Guided by the linearity properties of the fractional steps, we introduce a conditional-linear Fourier neural operator (CL-FNO) whose linear state mappings are modulated by nonlinear conditioning on inputs such as boundary geometry. We further introduce two complementary strategies, symmetry sampling and corrective cycling, to mitigate error accumulation during autoregressive rollout. Experiments on 2D and 3D fluid flows with varying geometries and initial conditions show that our full model reduces mean relative $L^2$ rollout error by 68\% in 2D and 40\% in 3D compared with a single full-step neural operator. At comparable error levels, neural-operator inference is approximately $10^3$ times faster per step than the Monte Carlo solver.
\end{abstract}

\section{Introduction}

Traditional numerical solvers for nonlinear dynamics in diverse geometries often decompose the system into operators, enabling faster predictions~\citep{macnamara2017operator}. Neural operators~\citep{li2021fourier} have been proposed as learning-based methods for solving complex PDEs by learning function-to-function mappings. However, most neural operators predict the next state with a single operator that must capture the full physical evolution, making generalization to unseen conditions and long-horizon rollout more challenging. Furthermore, existing operators do not respect the possible underlying structure of such an operator and do not fully exploit physical structure.

Fluid dynamics provides a natural setting for studying this issue, particularly in complex and diverse geometries, where both numerical solvers and learning-based surrogate models face distinct challenges~\citep{Sugimoto:2024:VelMCFluids,NEURIPS2023_70518ea4}. Numerical methods often require specialized meshing or boundary treatment, while learning-based surrogates must generalize across unseen geometries. 

Many traditional solvers for the Navier--Stokes equations solve the governing partial differential equations (PDEs) via operator splitting, in which a complex single time step is decomposed into advection, diffusion, and pressure projection~\citep{chorin1968numerical, Orszag_Kells_1980}. Advection and diffusion evolve the velocity field, while pressure projection enforces the divergence-free constraint on the velocity. For static geometries with homogeneous boundary conditions, only advection is nonlinear; diffusion and projection are linear in the evolving state. A monolithic next-step operator must rediscover all three from data, along with the linear structure of two of them.

The direct way to use this decomposition is to learn one operator per substep, supervised by the intermediate states the solver already produces. Table~\ref{tab:splitting} shows that this alone reduces rollout error by 10.7\% in 2D and 23.8\% in 3D over a single full-step operator. But the three substeps differ in what they demand of a learned model, and supervising them as three interchangeable regression problems leaves that difference unused.

\begin{figure*}
\centering
\resizebox{\linewidth}{!}{
\input{fig/pipeline}}
\caption{\textbf{Overview of learned operator splitting}. (a) Geometry-conditioned advection, diffusion, and projection operators advance the velocity field from frame $u^{(n)}$ to frame $u^{(n+1)}$. (b) Monte Carlo solver targets supervise each stage during training; no solves are required at inference. (c) Each operator uses a geometry encoder to generate gates that modulate spectral convolution blocks processing the input velocity field.}
\label{fig:pipeline}
\end{figure*}
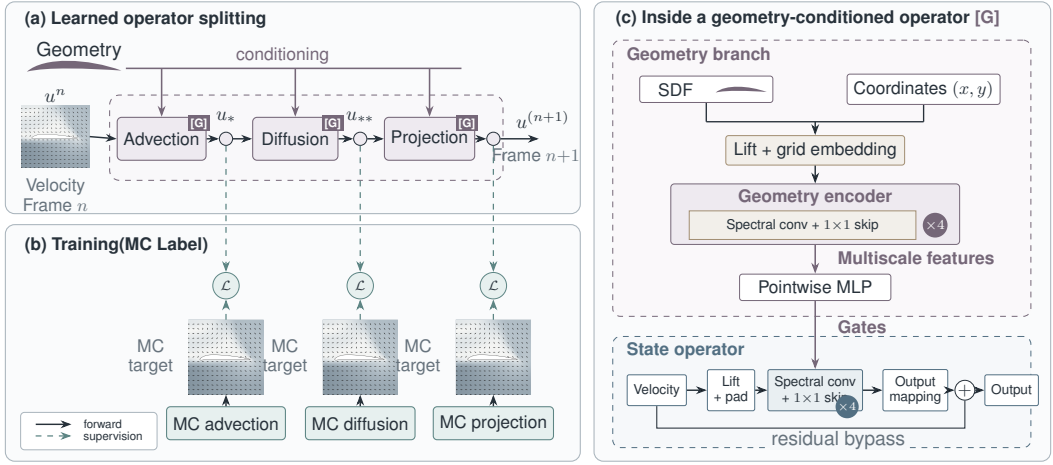

We therefore propose \textbf{neural operator splitting} as shown in Figure~\ref{fig:pipeline}, which learns each substep with an architecture matched to its structure. Its central component is a conditional-linear Fourier neural operator (CL-FNO), a two-branch design that separates a linear state-evolution branch from a nonlinear conditioning branch that incorporates geometry. For fixed geometry, the diffusion and projection operators are exactly linear in the evolving state, matching their true structure, while advection recovers its nonlinearity by receiving the state through the conditioning branch as well.

We further introduce two strategies for autoregressive rollout. Symmetry sampling resamples an element of the problem's spatial symmetry group at each operator application, disrupting the temporal coherence of prediction errors and  recovering most of the benefit of full group averaging at a small fraction of the cost. Corrective cycling trains an ordered sequence of operators each on states carrying the errors of its predecessors, without backpropagating through the rollout, and cycles through them during rollout to absorb distribution shift.

We evaluate on 2D and 3D incompressible flows with diverse geometries and initial conditions, testing generalization to unseen geometries and long rollout horizons. We primarily use a velocity-based Monte Carlo fluid solver~\citep{Sugimoto:2024:VelMCFluids} that handles varied geometries without body-fitted volume meshing and exposes the intermediate states needed for substep supervision. Additional spectral and Basilisk benchmarks demonstrate the framework's applicability to deterministic solvers.

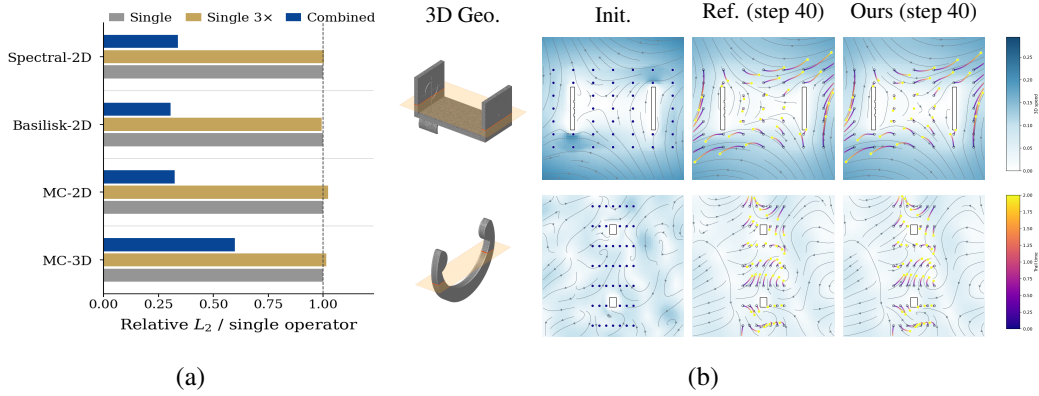
\begin{figure}[t]
  \centering
  \resizebox{\textwidth}{!}{%
  \begin{tikzpicture}[x=1cm,y=1cm]
    \node[anchor=north west,inner sep=0,outer sep=0] at (0,0.55)
      {\includegraphics[width=6cm,height=5.30cm]{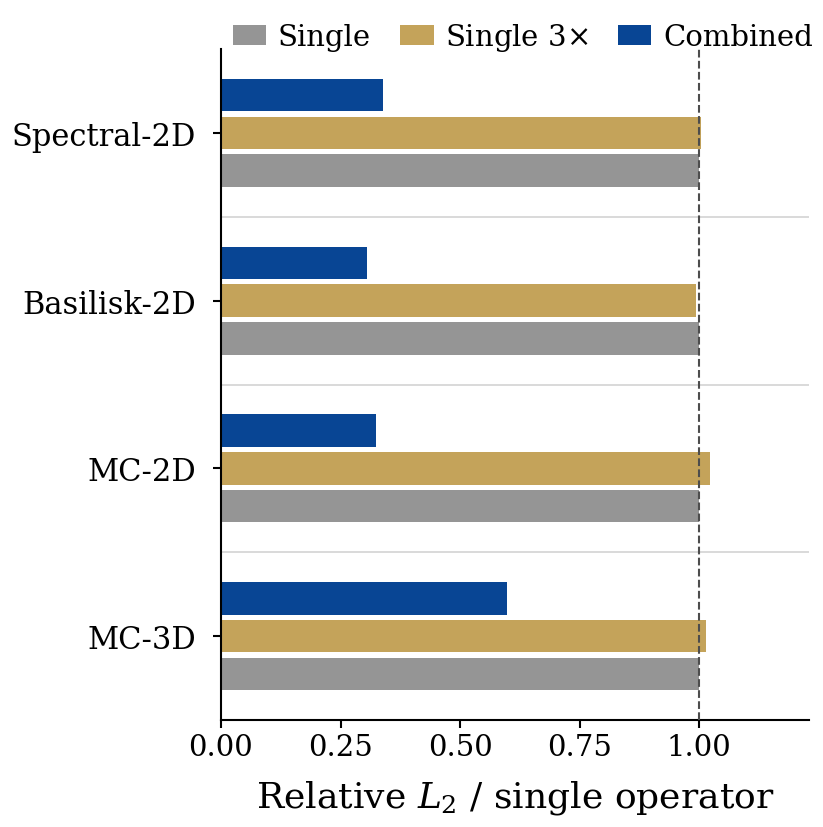}};
    \node[anchor=south] at (7.325,0.12) {3D Geo.};
    \node[anchor=south] at (9.715,0.12) {Init.};
    \node[anchor=south] at (12.105,0.12) {Ref. (step 40)};
    \node[anchor=south] at (14.495,0.12) {Ours (step 40)};
    \node[anchor=north west,inner sep=0,outer sep=0] at (6.20,0)
      {\includegraphics[width=2.25cm,height=2.25cm]{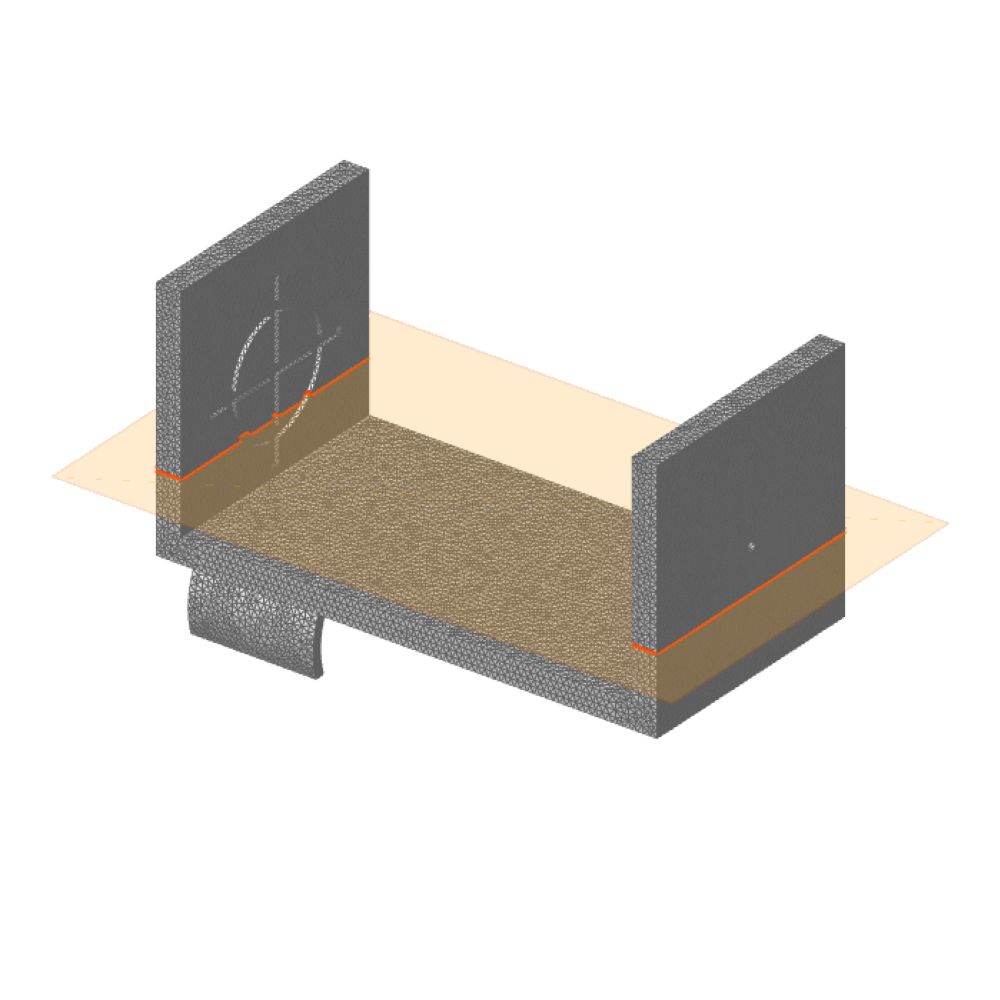}};
    \node[anchor=north west,inner sep=0,outer sep=0] at (8.59,0)
      {\includegraphics[width=2.25cm,height=2.25cm]{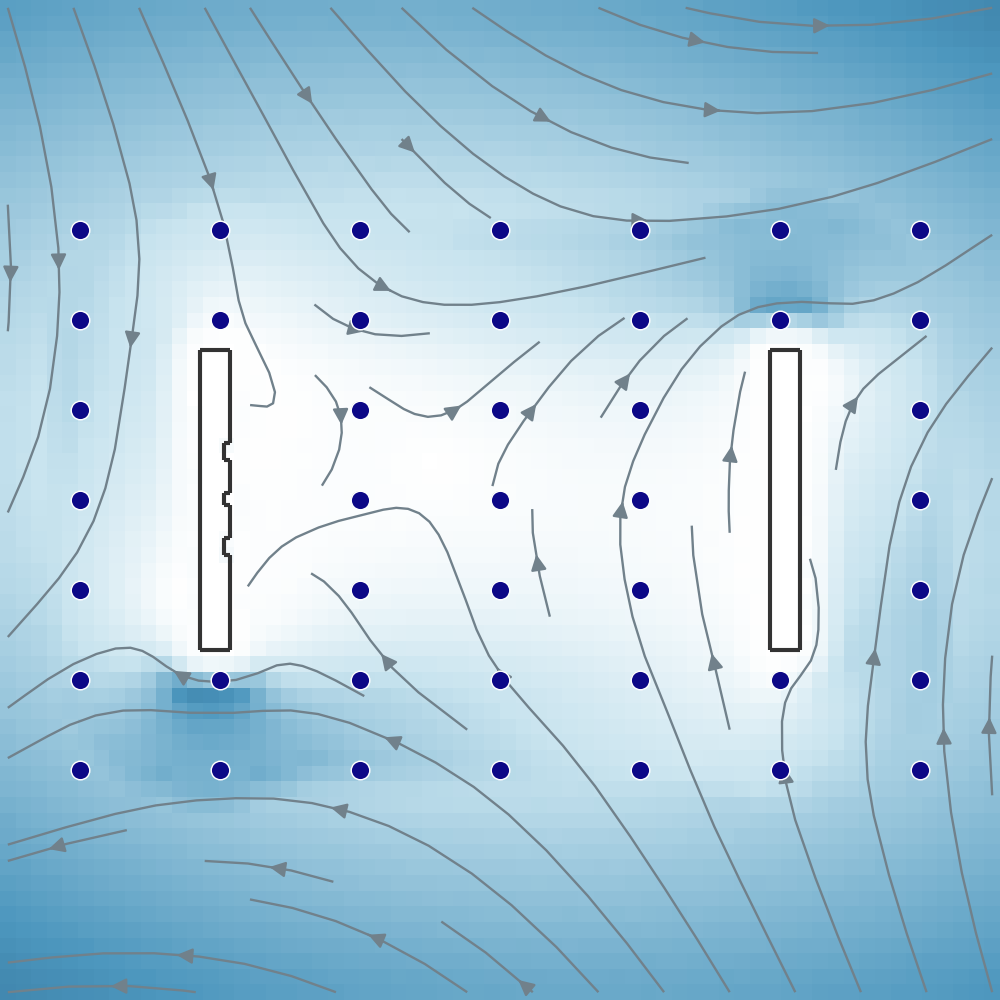}};
    \node[anchor=north west,inner sep=0,outer sep=0] at (10.98,0)
      {\includegraphics[width=2.25cm,height=2.25cm]{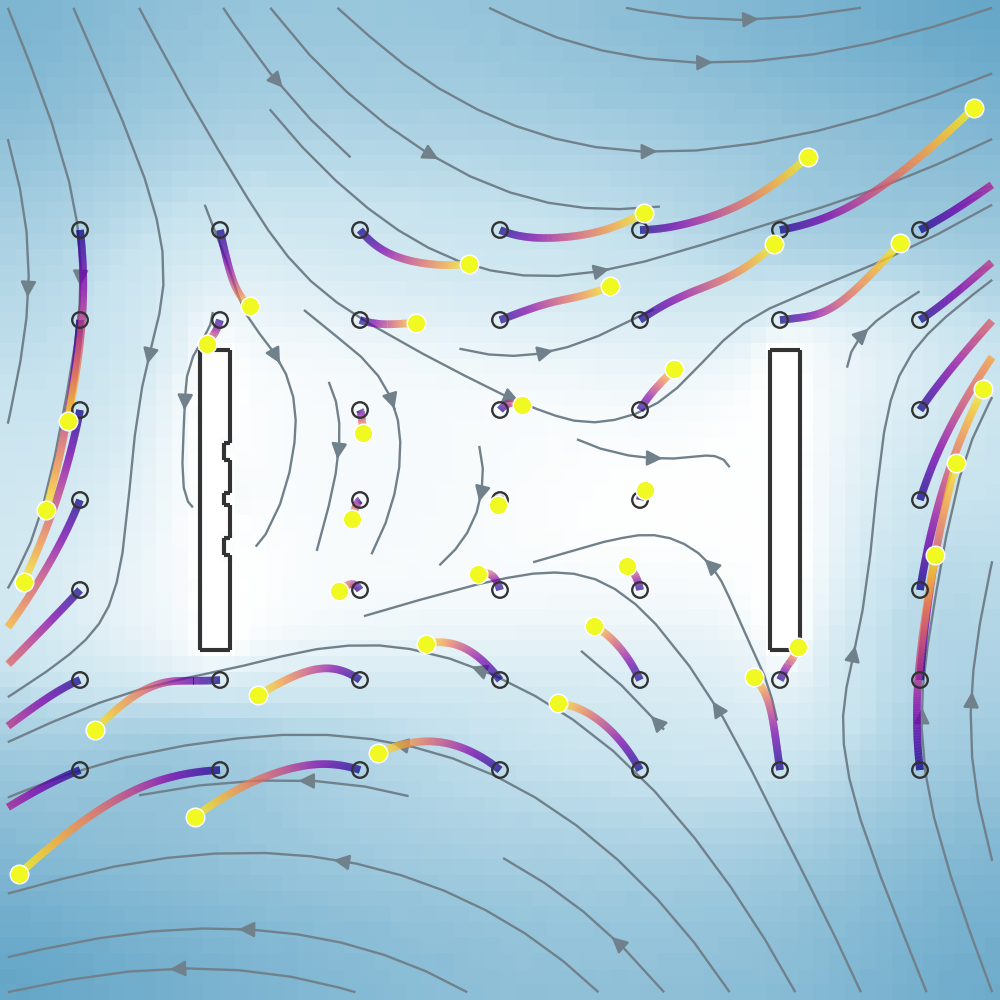}};
    \node[anchor=north west,inner sep=0,outer sep=0] at (13.37,0)
      {\includegraphics[width=2.25cm,height=2.25cm]{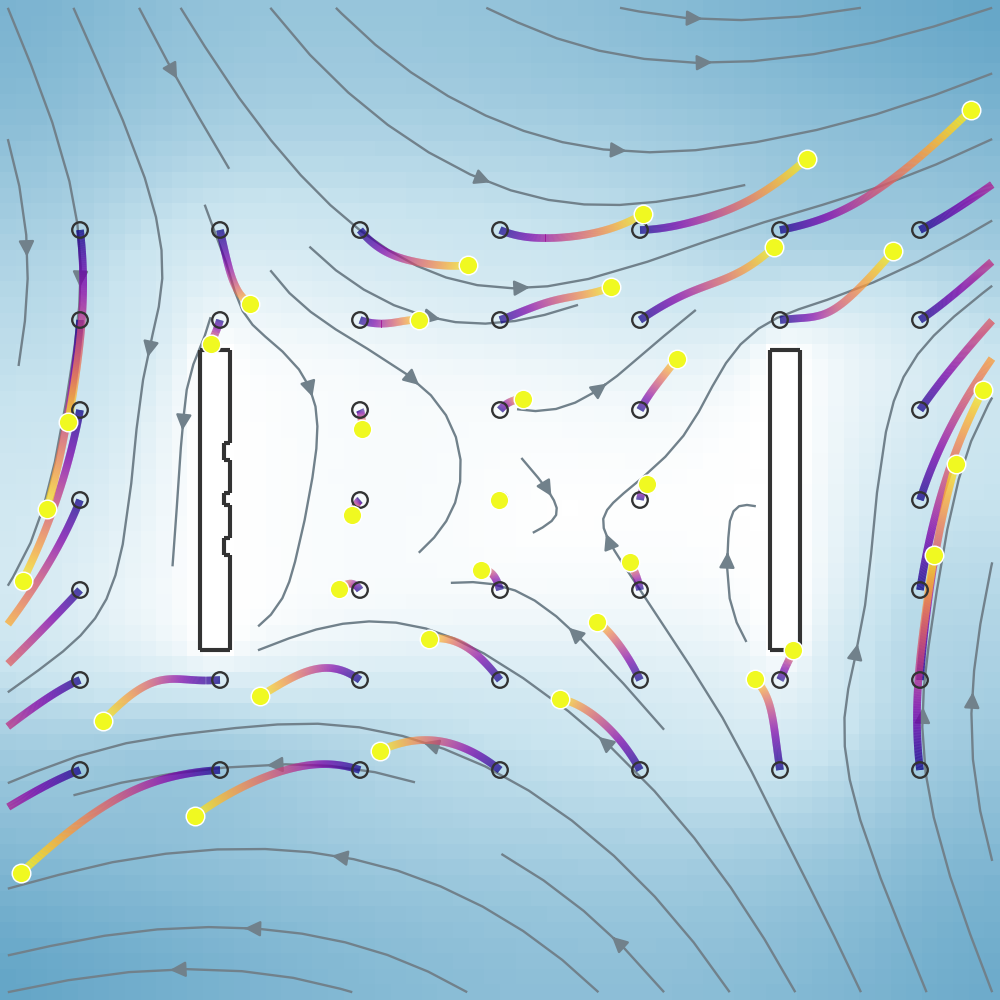}};
    \node[anchor=north west,inner sep=0,outer sep=0] at (6.20,-2.50)
      {\includegraphics[width=2.25cm,height=2.25cm]{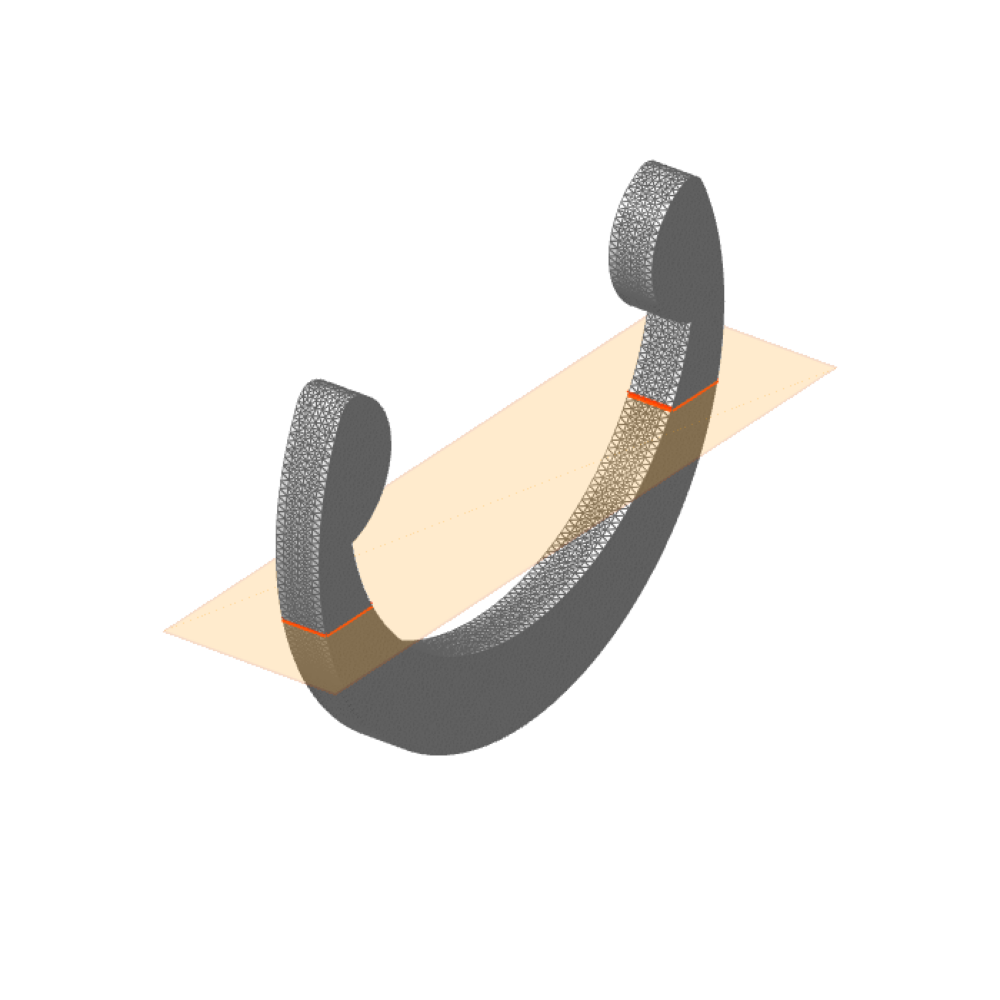}};
    \node[anchor=north west,inner sep=0,outer sep=0] at (8.59,-2.50)
      {\includegraphics[width=2.25cm,height=2.25cm]{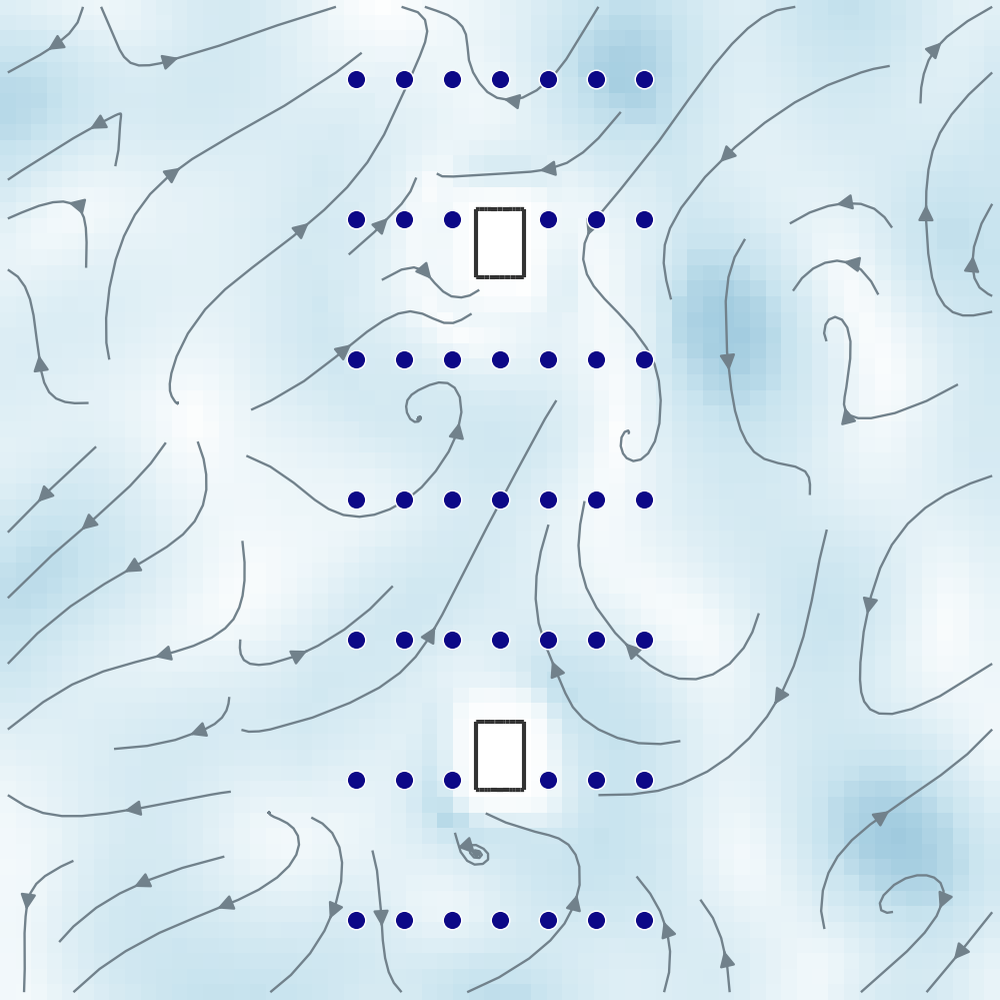}};
    \node[anchor=north west,inner sep=0,outer sep=0] at (10.98,-2.50)
      {\includegraphics[width=2.25cm,height=2.25cm]{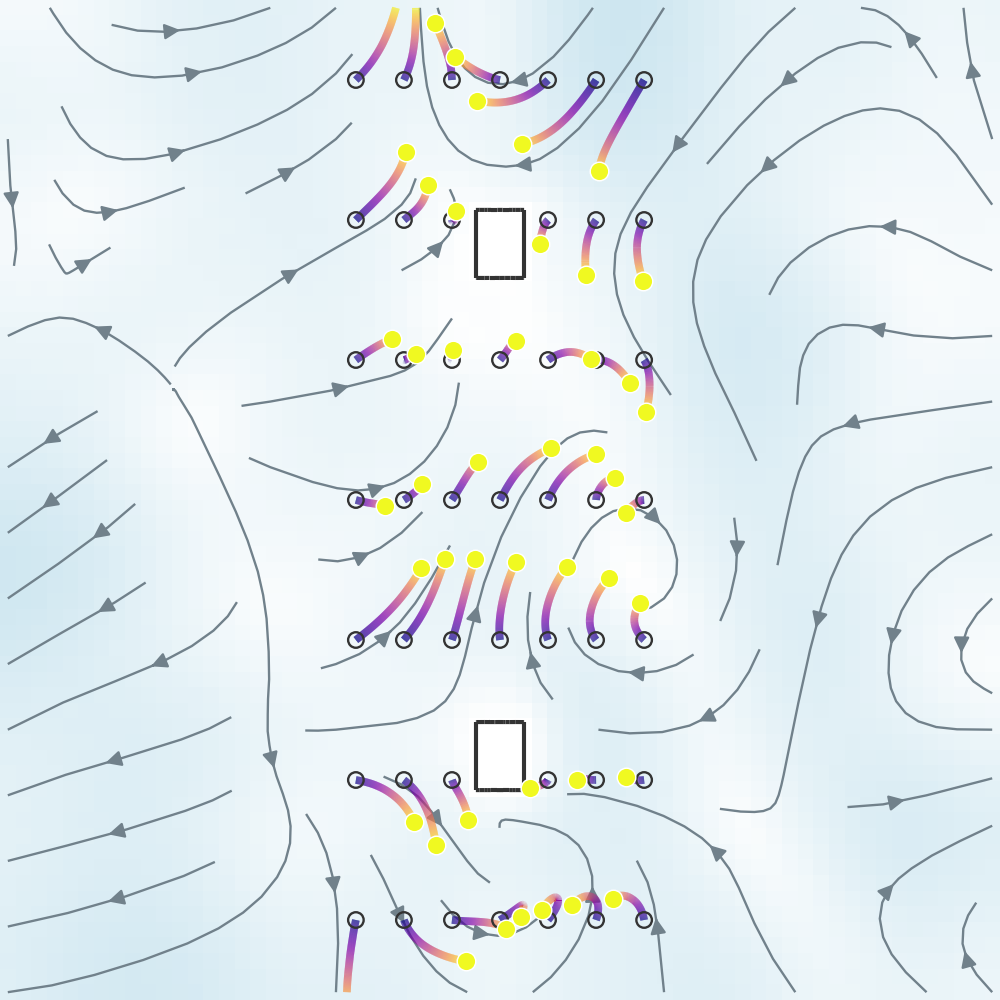}};
    \node[anchor=north west,inner sep=0,outer sep=0] at (13.37,-2.50)
      {\includegraphics[width=2.25cm,height=2.25cm]{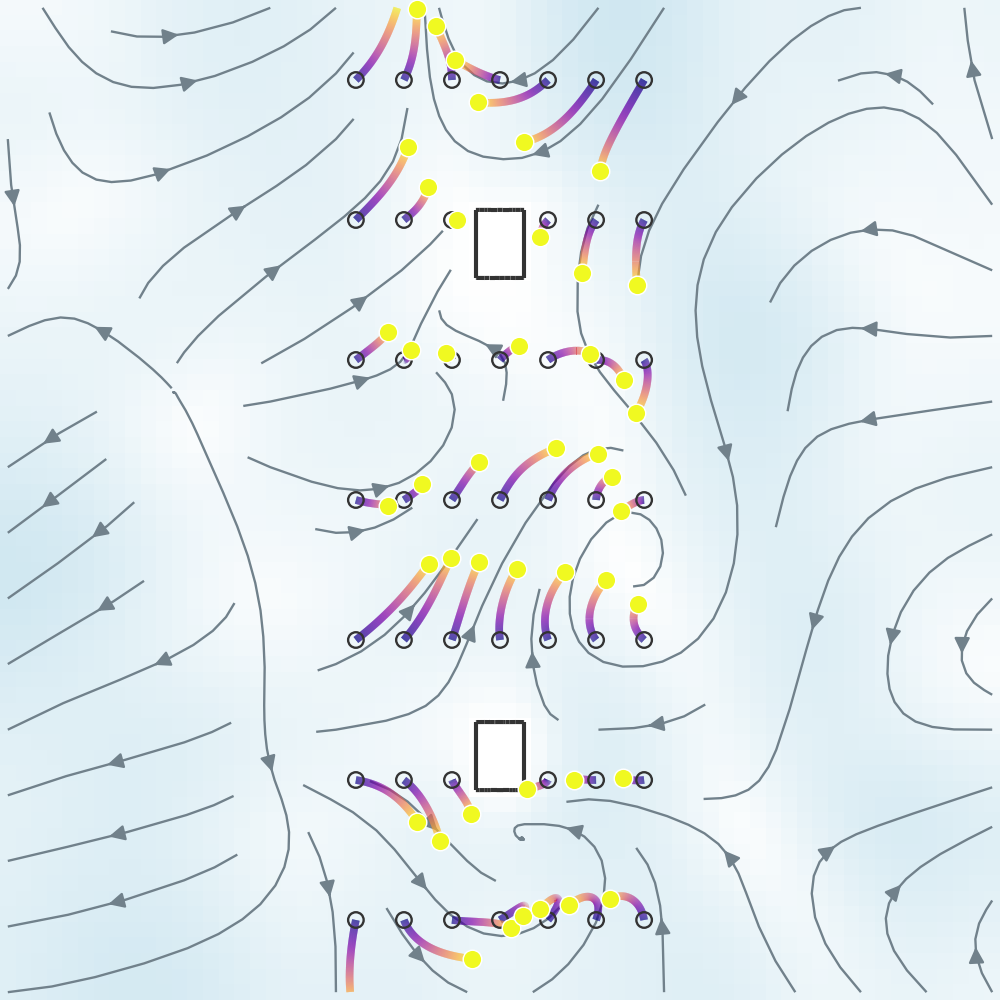}};
    \node[anchor=north west,inner sep=0] at (15.82,0.15)
      {\includegraphics[height=2.40cm]{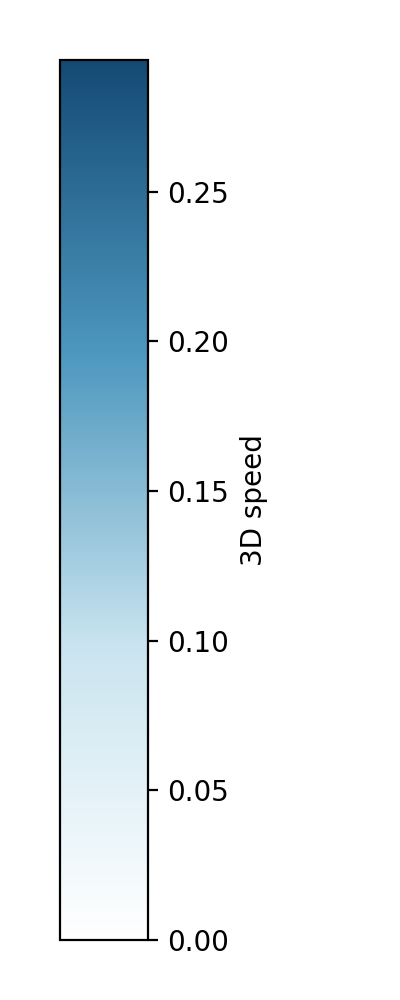}};
    \node[anchor=north west,inner sep=0] at (15.82,-2.35)
      {\includegraphics[height=2.40cm]{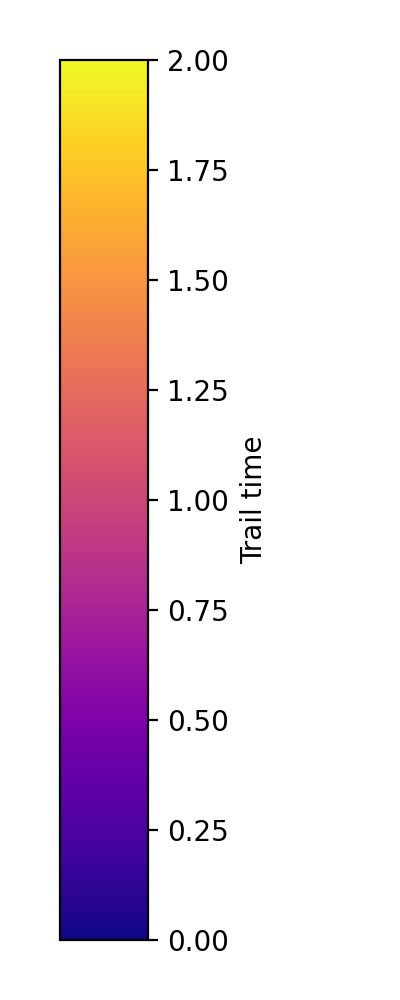}};
    \node[font=\large,anchor=north] at (3.0,-5.10) {(a)};
    \node[font=\large,anchor=north] at (11.15,-5.10) {(b)};
  \end{tikzpicture}%
  }
  \vspace{-2.0em}
  \caption{\textbf{(a) Improved accuracy}. We compare our neural operator splitting approach against a single operator with a matched network parameter count. Across four datasets, our method achieves a 2--3$\times$ reduction in relative $L^2$ error. \textbf{(b) 3D fluid rollout visualization}. We visualize the predicted 3D velocity field using particle trajectories on a representative slice of the domain. Across unseen geometries, our model produces rollout dynamics that remain spatially close to the reference solution.}
  \label{fig:placeholder}
\end{figure}

Our contributions include
\begin{itemize}[leftmargin=10pt]
\item \textbf{Solver-supervised fractional-step operators.} We learn the advection, diffusion, and pressure-projection substeps of a solver as separate neural operators, supervised directly by the intermediate states of the corresponding solver trajectory. On 2D and 3D flows with varying geometries, this reduces mean rollout error by 11\% and 24\% relative to a single full-step operator of the same architecture (Table~\ref{tab:splitting}).
\item \textbf{A conditional-linear operator architecture.} We separate the operator into a linear state-evolution branch and a nonlinear conditioning branch, allowing the learned map to remain exactly linear in the evolving state when the conditioning depends only on geometry. This matches the structure of diffusion and pressure projection, while leaving advection free to depend on the state. The design reduces rollout error by 18--61\% across settings (Table~\ref{tab:cl}).
\item \textbf{Symmetry sampling.} We resample a symmetry element from the problem's finite symmetry group at every operator application during rollout, reducing long-horizon error by 14--51\% without additional training (Table~\ref{tab:symmetry}). Stochastic sampling recovers nearly all of the benefit of full group averaging at a fraction of the forward passes (Fig.~\ref{fig:symmetry}), indicating that symmetry transformations help not only by enforcing equivariance but by disrupting the temporal coherence of prediction errors.
\item \textbf{Corrective cycling.} We train an ordered sequence of operators, each on states carrying the errors of its predecessors, and cycle through them during rollout. This addresses autoregressive distribution shift without backpropagating through a rollout, reducing error by up to 21\% (Table~\ref{tab:cc}).
\end{itemize}

\section{Related Works}
The Fourier neural operator (FNO)~\citep{li2021fourier} parameterizes global interactions in Fourier space and has been widely used for learning fluid dynamics, including Navier--Stokes flows. However, the standard FNO formulation is built around regular-grid discretizations, which makes complex and varying geometries challenging.

Existing approaches differ in how geometric information is incorporated into the operator. Geo-FNO~\citep{JMLR:v24:23-0064} learns a coordinate deformation from irregular domains to regular latent grids. DAFNO~\citep{NEURIPS2023_940a7634} incorporates a smoothed characteristic function of the domain into Fourier integral layers. GINO~\citep{NEURIPS2023_70518ea4} uses graph operators to couple irregular physical discretizations with a regular latent grid, where an FNO acts on the latent field. PCNO~\citep{ZENG2025118022} instead defines neural operators directly on point-cloud discretizations. These works primarily focus on geometric representation and generalization across domains, whereas our setting additionally emphasizes long-horizon fluid evolution across varying geometries.

Recent work has begun to explore operator splitting for decomposing complex PDE dynamics into subproblems. DOSnet~\citep{LAN2023112343} embeds classical operator splitting directly into a neural architecture composed of multiple learnable operator-splitting blocks. Within neural operators, \citeauthor{pmlr-v300-gopakumar26a} decompose PDE dynamics into learned and numerically discretized physical operators whose contributions are combined before time integration. \citeauthor{sun2026origo} uses a context-conditioned Strang-splitting backbone to combine learned global and local mechanisms for multiphysics pretraining. Another approach pretrains operators for elementary or single-physics dynamics and subsequently combines them for multiphysics systems through a learned adaptation block~\citep{hmida2026compositional}. \citeauthor{serrano2026disco} instead compose operators from a pretrained dictionary through test-time operator splitting. In contrast, our neural operator splitting follows the solver's fractional-step sequence and directly supervises each suboperator with consecutive intermediate states from the same solver trajectory.

A separate line of work has explored operator architectures that distinguish different input roles and structure their interactions accordingly. VarMiON~\citep{56b77649b60443079266e65254c18cd0} derives this structure from variational discretizations. Neural Green's Operators~\citep{cf62894765ac45ccbdc080450ad4aef3} and Neural Green's Functions~\citep{NEURIPS2025_448abd48} use Green-function formulations to preserve linear dependence on forcing and boundary data, while learning how the solution operator varies with PDE coefficients and domain geometry, respectively. Separately, HyperFNO~\citep{alesiani2022hyperfno} generates FNO parameters from physical PDE parameters, while \citeauthor{ICLR2026_c253f6fe} generate parameter-dependent shifts for individual FNO layers. Our conditional-linear architecture uses FNO-generated layer-wise gates to condition the operator while preserving linearity in the evolving state.

Symmetry has been incorporated into neural PDE solvers through equivariant architectures and group-based symmetrization. G-FNO~\citep{pmlr-v202-helwig23a} builds rotation and reflection equivariance directly into Fourier neural operator layers. Beyond architectural constraints, frame averaging~\citep{puny2022frame} averages predictions over input-dependent frames, while probabilistic symmetrization~\citep{NEURIPS2023_3b5c7c9c} learns an input-conditioned distribution over group transformations. \citeauthor{DBLP:journals/corr/abs-2511-09573} apply test-time group averaging to autoregressive physical surrogates, using exact averaging for finite groups and Monte Carlo estimates for continuous symmetries. Our approach uses single-sample symmetrization for finite groups and refreshes the sampled realization across operator applications during rollout to mitigate coherent error accumulation.

Finally, autoregressive neural PDE solvers face distribution shift as model predictions are recursively fed back as inputs. Pushforward training~\citep{DBLP:conf/iclr/BrandstetterWW22} mitigates this shift by training on self-generated states. Our corrective cycling instead distributes adaptation to rollout-induced states across a sequence of operator instances that are cycled during rollout.

\section{Neural Operator Splitting}
In this section, we introduce the Navier--Stokes problem setup, our neural operator splitting framework and a conditional linear architecture motivated by the structure of the split operators, and associated techniques for improving long-horizon rollout performance.

\subsection{Neural Operator Splitting of the Navier–Stokes Equations }

We consider incompressible fluid evolution governed by
\begin{equation}
\frac{\partial \mathbf{u}}{\partial t} +(\mathbf{u}\cdot\nabla)\mathbf{u}
= -\frac{1}{\rho}\nabla p+\nu\nabla^2\mathbf{u},
\qquad \nabla\cdot\mathbf{u}=0.
\end{equation}
Here \(\mathbf{u}\) is the velocity field, \(p\) is the pressure, \(t\) is time, \(\rho\) is the fluid density, and \(\nu\) is the kinematic viscosity. The operators \(\nabla\), \(\nabla\cdot\), and \(\nabla^2\) denote the spatial gradient, divergence, and Laplacian, respectively.

We use the operator splitting induced by the fractional-step solver, which decomposes one solver step into the advection, diffusion, and pressure-projection substeps:
\begin{align}
\mathbf{u}^{n+1}_{*} &= A(\mathbf{u}^{n}), 
&& \partial_\tau \mathbf{u} + (\mathbf{u}\cdot\nabla)\mathbf{u}=0,\quad \tau\in[0,\Delta t], \\
\mathbf{u}^{n+1}_{**} &= D(\mathbf{u}^{n+1}_{*}), 
&& \partial_\tau \mathbf{u} = \nu\nabla^2 \mathbf{u},\quad \tau\in[0,\Delta t], \\
\mathbf{u}^{n+1} &= P(\mathbf{u}^{n+1}_{**}), 
&& \nabla^2 \overline{p} = \nabla\cdot \mathbf{u}^{n+1}_{**},\quad
\mathbf{u}^{n+1}=\mathbf{u}^{n+1}_{**}-\nabla\overline{p}.
\end{align}
Here \(n\) indexes the solver time steps, \(\mathbf{u}^{n}\) denotes the velocity at time \(t^n\), \(\Delta t=t^{n+1}-t^n\) is the time-step size, and \(\tau\) is the local time within each substep. The fields \(\mathbf{u}^{n+1}_{*}\) and \(\mathbf{u}^{n+1}_{**}\) are the intermediate velocities after the advection and diffusion substeps, respectively, and \(A\), \(D\), and \(P\) denote the corresponding physical substep operators. The scaled pressure is \(\overline p=\frac{\Delta t}{\rho}p\).

On solid boundaries, the velocity satisfies
\begin{equation}
\mathbf{u}=\mathbf{u}_s,
\end{equation}
where \(\mathbf{u}_s\) denotes the prescribed solid-boundary velocity, with \(\mathbf{u}_s=\mathbf{0}\) for stationary boundaries. Here \(\overline p\) satisfies the Neumann condition
\begin{equation}
\frac{\partial \overline p}{\partial n}
=
\mathbf n\cdot
\left(\mathbf u^{n+1}_{**}-\mathbf u_s\right),
\end{equation}
where \(\mathbf n\) is the outward unit normal from the fluid domain and \(\partial/\partial n=\mathbf n\cdot\nabla\) denotes the normal derivative.

Equivalently,
\begin{equation}
\mathbf{u}^{n} \xrightarrow{A} \mathbf{u}^{n+1}_{*}
\xrightarrow{D} \mathbf{u}^{n+1}_{**}
\xrightarrow{P} \mathbf{u}^{n+1}.
\end{equation}
The solver provides \(\mathbf{u}^n\), \(\mathbf{u}^{n+1}_{*}\), \(\mathbf{u}^{n+1}_{**}\), and \(\mathbf{u}^{n+1}\). We therefore learn
three neural operators: \(\mathcal{F}_A\) for the advection map \(A\), \(\mathcal{F}_D\) for the diffusion map \(D\), and \(\mathcal{F}_P\) for the pressure-projection map \(P\). Each operator is trained independently using direct supervision from the corresponding solver input--output pair, with each operator learning a single physical substep. At inference, they are applied in sequence,
\begin{equation}
\hat{\mathbf{u}}^{n+1}
=
\mathcal{F}_P\circ \mathcal{F}_D\circ \mathcal{F}_A
[\hat{\mathbf{u}}^n],
\end{equation}
where \(\hat{\mathbf{u}}^n\) denotes the predicted velocity at time step \(n\), and \(\circ\) denotes operator composition.

This setup enables a direct comparison between operator splitting and a single neural operator trained directly on the full-step map \(\mathbf{u}^n\mapsto\mathbf{u}^{n+1}\) in autoregressive rollout.

\subsection{Conditional Linear Architecture}

We separate the neural operator into a linear branch for the evolving state \(\mathbf{u}\) and a nonlinear  branch for the conditioning input \(\mathbf{c}\). The conditioning branch processes \(\mathbf{c}\) with an FNO and collects the feature fields after lifting and after each FNO layer. These features are fused pointwise by an MLP into a shared conditioning representation. For each layer \(\ell\)  of the state branch, separate heads then produce input and output gates,
\begin{equation}
\Gamma_\ell^\mathrm{in}(\mathbf{c}),
\qquad
\Gamma_\ell^\mathrm{out}(\mathbf{c}).
\end{equation}
These gates modulate the state features before and after the corresponding linear FNO block, so the conditioning branch affects the state evolution only through multiplicative modulation.

Let \(\mathbf{z}_\ell\) denote the input feature to layer \(\ell\) of the state branch, and define its linear FNO block as
\begin{equation}
L_\ell(\mathbf{z}_\ell)
=
K_\ell(\mathbf{z}_\ell)+W_\ell(\mathbf{z}_\ell),
\end{equation}
where \(K_\ell\) is a bias-free spectral convolution and \(W_\ell\) is a pointwise linear map. The corresponding state update is
\begin{equation}
\mathbf{z}_{\ell+1}
=
\Gamma_{\ell}^\mathrm{out}(\mathbf{c})
\odot
L_\ell\!\left(
\Gamma_{\ell}^\mathrm{in}(\mathbf{c})
\odot \mathbf{z}_\ell
\right),
\end{equation}
where \(\odot\) denotes pointwise multiplication.

Therefore, for fixed \(\mathbf{c}\), the resulting neural operator \(\mathcal{F}(\mathbf{u};\mathbf{c})\) is linear in the evolving state \(\mathbf{u}\).

This design reflects the operator structure of the split dynamics. Diffusion,
\begin{equation}
\partial_\tau \mathbf{u}=\nu\nabla^2\mathbf{u},
\end{equation}
and pressure projection,
\begin{equation}
\nabla^2\overline p=\nabla\cdot\mathbf{u},
\qquad
P(\mathbf{u})=\mathbf{u}-\nabla\overline p,
\end{equation}
are linear in \(\mathbf{u}\) for fixed geometry. Their conditioning branches therefore depend only on geometric representations and spatial coordinates.

Advection is nonlinear overall,
\begin{equation}
\partial_\tau \mathbf{u} + (\mathbf{u}\cdot\nabla)\mathbf{u}=0,
\end{equation}
but is linear in the advected field when the advecting velocity is held fixed. In self-advection, \(\mathbf{u}\) serves as both the advected field and the advecting velocity, so we additionally include \(\mathbf{u}\) in the conditioning input \(\mathbf{c}\). Since advection is the only nonlinear component of the split dynamics considered here, the same state-dependent conditioning is used for the single-step operator.

\subsection{Symmetry Sampling}

We consider finite spatial symmetry groups that preserve the governing equations under the corresponding transformation of the spatial domain and vector-valued flow fields. Specifically, we use the dihedral group \(D_4\), consisting of the rotations and reflections of the square, for the 2D problems, and the full octahedral group \(O_h\), consisting of the rotational and reflection symmetries of the cube, for the 3D problem. We denote either group by \(G\), and let \(\mathcal{T}_g\) denote the action of \(g\in G\) on spatially defined inputs and outputs, including the corresponding transformation of vector components for vector-valued fields.

For a learned operator \(\mathcal{F}\), its symmetry-transformed operator under \(g\) is
\begin{equation}
\mathcal{F}^{(g)}
=
\mathcal{T}_g^{-1}\circ \mathcal F\circ \mathcal{T}_g ,
\end{equation}
where \(\mathcal{T}_g^{-1}\) denotes the inverse symmetry action. 

Rather than using the same symmetry transform throughout a rollout, we independently sample
\begin{equation}
g_k\sim \mathrm{Unif}(G)
\end{equation}
at each operator application \(k\), where \(\mathrm{Unif}(G)\) denotes the uniform distribution over \(G\). At each application, the sampled \(g_k\) is used to form the corresponding transformed version of the learned operator.

Beyond the commonly discussed train--rollout distribution shift, autoregressive prediction can also amplify correlated approximation errors, as repeatedly applying the same operator realization to successive states may produce spatially aligned errors that accumulate coherently over time. Random symmetry sampling varies the symmetry transform across applications and can mitigate this coherent error accumulation.

\subsection{Corrective Cycling}

Corrective cycling uses an ordered sequence of \(M\) learned operators, \((\mathcal{F}_1,\ldots,\mathcal{F}_M)\). Let \(\mathbf u_k\) and \(\hat{\mathbf u}_k\) denote the ground-truth and predicted states at operator application \(k\), respectively. Starting from \(\hat{\mathbf u}_k=\mathbf u_k\), we recursively define
\begin{equation}
\hat{\mathbf u}_{k+j}
=
\mathcal{F}_j\!\left(\hat{\mathbf u}_{k+j-1}\right),
\qquad j=1,\ldots,M.
\end{equation}

The operators are trained sequentially. After \(\mathcal{F}_1,\ldots,\mathcal{F}_{j-1}\) have been trained and frozen, the training inputs for \(\mathcal{F}_j\) are generated offline by recursively applying these preceding operators starting from ground-truth states. Thus, \(\mathcal{F}_j\) is trained to predict \(\mathbf u_{k+j}\) from \(\hat{\mathbf u}_{k+j-1}\),  which
carries their accumulated errors.

During rollout, the operators are applied cyclically,
\begin{equation}
\mathcal{F}_1\rightarrow \mathcal{F}_2\rightarrow\cdots\rightarrow \mathcal{F}_M
\rightarrow \mathcal{F}_1\rightarrow\cdots .
\end{equation}
For split models, the corrective-cycle index advances after each substep application, continuing across advection, diffusion and pressure projection rather than restarting at each solver time step.

\section{Experiment Design and Results}
\label{sec:MC}
In this section, we explain our experiment design for our method as well as final results.
\subsection{Datasets}
We use four fluid benchmarks for training and evaluation: Spectral-2D, Basilisk-2D, MC-2D, and MC-3D. We use VelMCFluids~\citep{Sugimoto:2024:VelMCFluids} as our primary data-generation solver because it handles varied solid geometries through ray--boundary intersection queries without requiring a body-fitted volume mesh, while allowing velocity fields to be stored on the regular grids used by our neural operators. Its explicit advection, diffusion, and pressure-projection stages provide intermediate states that directly match our supervision targets. By comparison, standard PISO/PIMPLE-based OpenFOAM solvers combine convection and viscous terms within a momentum equation and enforce pressure--velocity coupling through correction loops~\citep{weller1998tensorial,openfoamSolutionControl}, so obtaining the same three physical substep targets would require additional adaptation of the numerical scheme. These properties make VelMCFluids convenient for our experiments, but the learning framework does not require stochastic supervision. To demonstrate its applicability to deterministic solvers, we additionally implement a spectral solver with separate advection, diffusion, and pressure-projection stages and evaluate a modified Basilisk solver~\citep{POPINET20095838}. The resulting Spectral-2D and Basilisk-2D benchmarks test the framework without solid geometry and with embedded obstacles, respectively, complementing the MC benchmarks with deterministic evaluations. Full dataset and solver details are provided in Appendix~\ref{app:data}.

Spectral-2D uses standard held-out trajectories, while the other three benchmarks evaluate zero-shot generalization to unseen geometries. Basilisk-2D is trained on circular obstacles and tested on squares; MC-2D and MC-3D are tested on unseen NACA airfoils and ABC geometries~\citep{Koch_2019_CVPR}, respectively. Spectral-2D, Basilisk-2D, and MC-2D are trained on 100-step trajectories and evaluated with 200-step rollouts, additionally testing temporal extrapolation. Due to its higher computational cost, MC-3D uses 50 steps for both training and evaluation.

\subsection{Training and Evaluation Protocol}
All learned operators predict residual velocity updates,
\begin{equation}
\mathcal{F}(\mathbf{u};\mathbf{c})
=
\mathbf{u}
+
\mathcal{R}(\mathbf{u};\mathbf{c}),
\end{equation}
where $\mathbf{u}$ is the input state, and $\mathbf{c}$ denotes the physical and geometric conditioning variables. For the single-operator model, these are consecutive full-step states; for Neural Operator Splitting, the same formulation is applied to each supervised fractional step pair. All operators are trained with relative \(L^2\) loss over fluid grid points. Normalization, padding, data augmentation, model architecture, and other training details are provided in Appendix~\ref{app:setup}.

We evaluate autoregressive rollouts over the fluid region using pooled relative \(L^2\) rollout error, referred to as rollout error below. Tables pool squared errors and reference energies across trajectories and time; curves pool them across trajectories at each step. For MC benchmarks, tables report variance-corrected errors, while rollout curves show standard errors together with MC noise; deterministic benchmarks use standard errors throughout. Appendix~\ref{app:variance-corrected-error} defines both error metrics and the MC noise estimates. Additional evaluation metrics are provided in Appendix~\ref{app:additional-metrics}.

Figure~\ref{fig:combined_rollout} shows representative 2D and 3D rollouts of the single-operator baseline and the full model.

\begin{figure}[t]
    \centering
    \includegraphics[width=\linewidth]{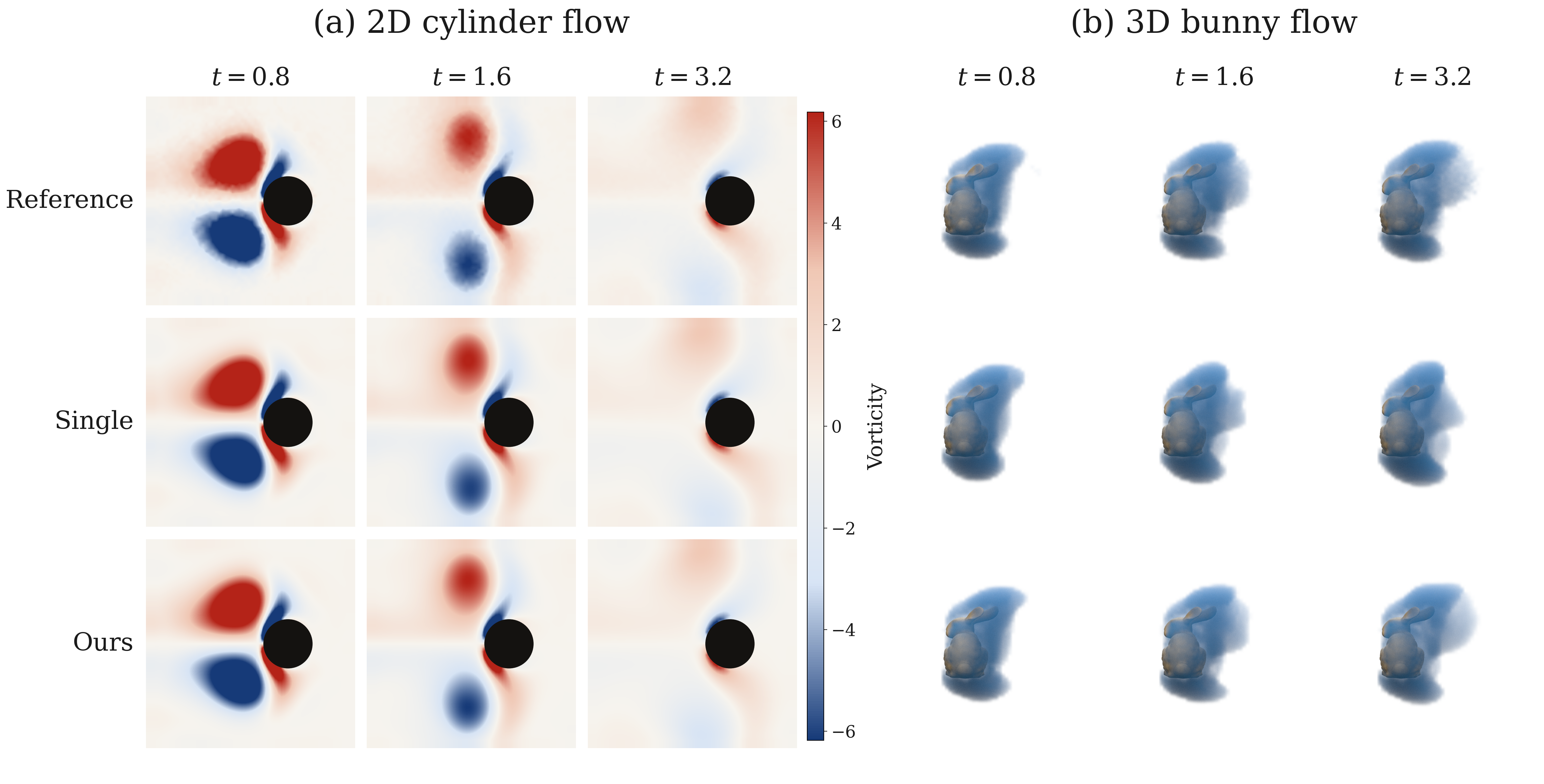}
    \caption{\textbf{Qualitative rollout comparison in 2D and 3D.}
    Rows show the MC reference, a single full-step operator (Single), and our combined model (Ours). Ours uses splitting, CL, SS and CC. Columns show $t=0.8$, $1.6$, and $3.2$.
    \textbf{(a)} Flow around a circular obstacle, colored by signed vorticity with a shared scale.
    \textbf{(b)} Flow around a bunny-shaped obstacle. Blue volumes show vortex regions identified by the $\lambda_2$ criterion. All 3D panels use the same rendering parameters.}
    \label{fig:combined_rollout}
\end{figure}




\subsection{Neural Operator Splitting}
For both deterministic and Monte Carlo benchmarks, we generate training trajectories and record the velocity fields after each advection, diffusion, and projection substep. The single operator is supervised by the full-step pairs \((\mathbf u^n,\mathbf u^{n+1})\), while the split operators are independently supervised by the corresponding substep pairs \((\mathbf u^n,\mathbf u_*^n)\), \((\mathbf u_*^n,\mathbf u_{**}^n)\), and \((\mathbf u_{**}^n,\mathbf u^{n+1})\). At inference time, the split operators are applied sequentially.

Although splitting does not improve performance in the simpler setting without geometry, it consistently outperforms the single operator for fluid dynamics with geometry (Table~\ref{tab:splitting}), with lower errors both within the training horizon and in temporal extrapolation in Fig.~\ref{fig:cl}. To control for the increased parameter count introduced by using three split operators, we additionally train single-operator baselines with three times the number of FNO layers, yielding a comparable parameter count. As shown in Table~\ref{tab:splitting}, this threefold increase in model size provides essentially no performance improvement, indicating that the gains from splitting do not arise from increased model capacity. 

\begin{figure}[t]
    \centering
    \includegraphics[width=1\linewidth]{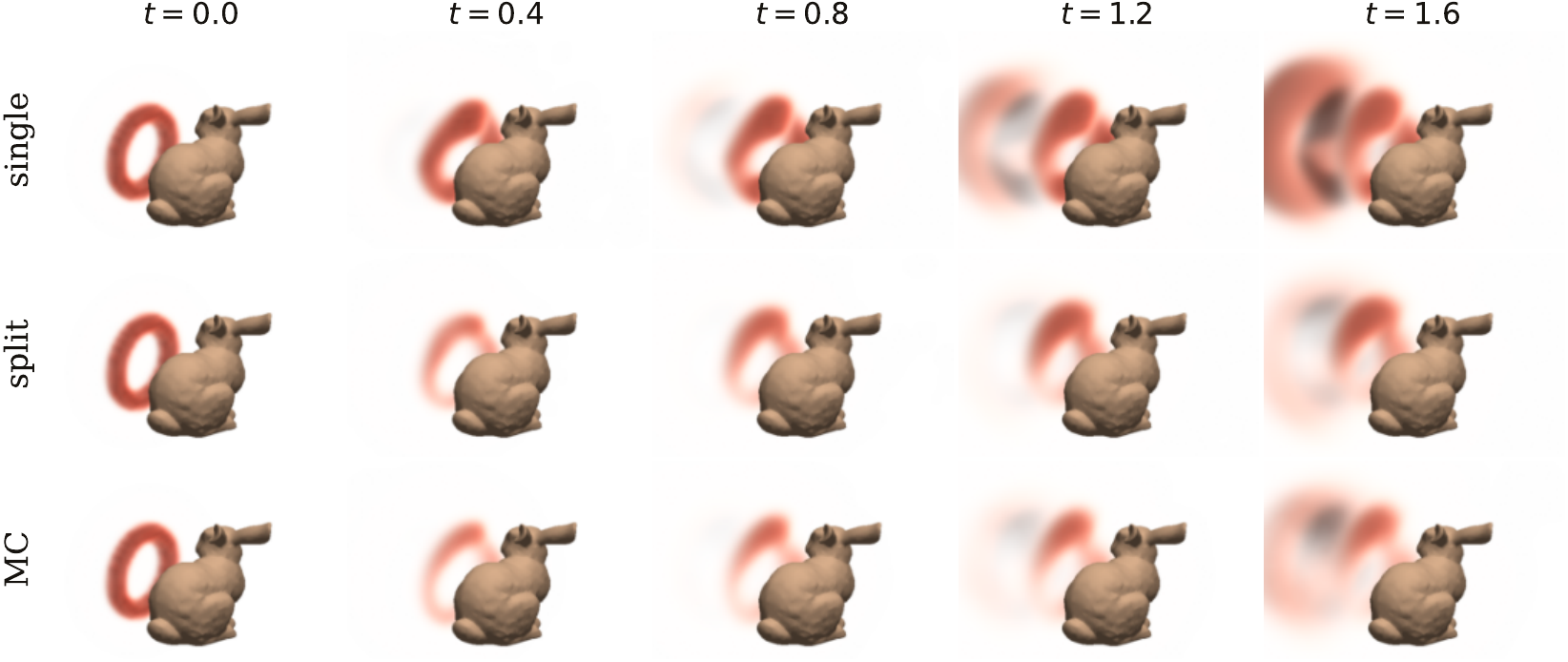}
    \caption{\textbf{Vortex-ring interaction with a bunny-shaped obstacle.} Rows show a single full-step operator (single), an operator-splitting model (split), and the MC reference. Columns show $t=0.0$, $0.4$, $0.8$, $1.2$, and $1.6$. Red volumes show the magnitude of azimuthal vorticity; brown surfaces show the solid obstacle.}
    \label{fig:bunny_ring}
\end{figure}

A qualitative example is the vortex ring impinging on the bunny, where the evolution of the vortex shape is visibly closer to the MC reference for the split model than for the single operator, as shown in Fig.~\ref{fig:bunny_ring}.

\begin{table}[t]
\centering
\caption{Relative $L^2$ rollout error of single-operator and split models. Single $3\times$ uses three times as many FNO layers as Single. Percentages are relative to Single.}
\label{tab:splitting}
\begin{tabular}{lccc}
\toprule
Dataset & Single & Single 3$\times$ & Split \\
\midrule
Spectral-2D & $\mathbf{1.626\times10^{-3}}$ & $1.634\times10^{-3}$ ($+$0.5\%) & $2.630\times10^{-3}$ ($+$61.8\%) \\
Basilisk-2D & $5.286\times10^{-2}$ & $5.251\times10^{-2}$ ($-$0.7\%) & $\mathbf{4.919\times10^{-2}}$ ($-$6.9\%) \\
MC-2D & $7.78\times10^{-2}$ & $7.96\times10^{-2}$ ($+$2.2\%) & $\mathbf{6.95\times10^{-2}}$ ($-$10.7\%) \\
MC-3D & $5.07\times10^{-2}$ & $5.14\times10^{-2}$ ($+$1.3\%) & $\mathbf{3.86\times10^{-2}}$ ($-$23.8\%) \\
\bottomrule
\end{tabular}
\end{table}

\subsection{Conditional Linear Architecture}

\begin{figure}[t]
    \centering
    \includegraphics[width=1\linewidth]{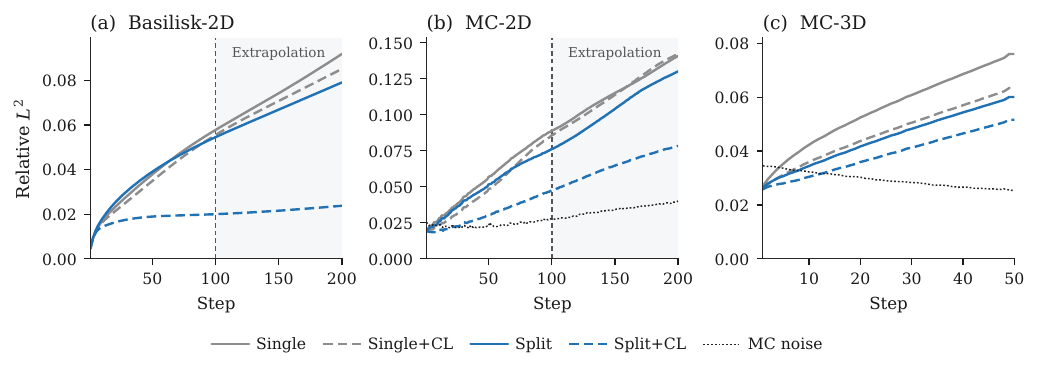}
     \caption{Relative $L^2$ rollout error of standard and conditional-linear (CL) single-operator and split models. Black dotted curves show MC noise; shaded regions mark extrapolation beyond the training horizon.}
    \label{fig:cl}
\end{figure}

For geometry-dependent datasets, geometry and physical coordinates are provided only to the conditioning branch. For the single and advection operators, velocity enters both branches, whereas for diffusion and projection it enters only the linear branch. Spectral-2D uses neither geometry nor coordinate inputs; its diffusion and projection operators therefore reduce to the linear FNO. As shown in Fig.~\ref{fig:cl} and Table~\ref{tab:cl}, this conditional linear (CL) design reduces long-horizon rollout error for the split models across all benchmarks, whereas the benefit for single models depends on the dataset.

\begin{table}[t]
\centering
\caption{ Relative $L^2$ rollout error of the standard and conditional-linear (CL) single and split models. Percentages indicate the relative change from the corresponding standard model. }
\label{tab:cl}
\small
\begin{tabular}{lcccc}
\toprule
Dataset & Standard Single & CL Single & Standard Split & CL Split \\
\midrule
Spectral-2D & $1.626\times10^{-3}$ & $2.416\times10^{-3}$ ($+$48.6\%) & $2.630\times10^{-3}$ & $\mathbf{1.031\times10^{-3}}$ ($-$60.8\%) \\
Basilisk-2D & $5.286\times10^{-2}$ & $4.999\times10^{-2}$ ($-$5.4\%) & $4.919\times10^{-2}$ & $\mathbf{1.901\times10^{-2}}$ ($-$61.3\%) \\
MC-2D & $7.78\times10^{-2}$ & $7.59\times10^{-2}$ ($-$2.5\%) & $6.95\times10^{-2}$ & $\mathbf{3.97\times10^{-2}}$ ($-$42.9\%) \\
MC-3D & $5.07\times10^{-2}$ & $4.09\times10^{-2}$ ($-$19.3\%) & $3.86\times10^{-2}$ & $\mathbf{3.17\times10^{-2}}$ ($-$17.9\%) \\
\bottomrule
\end{tabular}
\end{table}

\subsection{Symmetry Sampling}

\begin{figure}[t]
    \centering
    \includegraphics[width=1\linewidth]{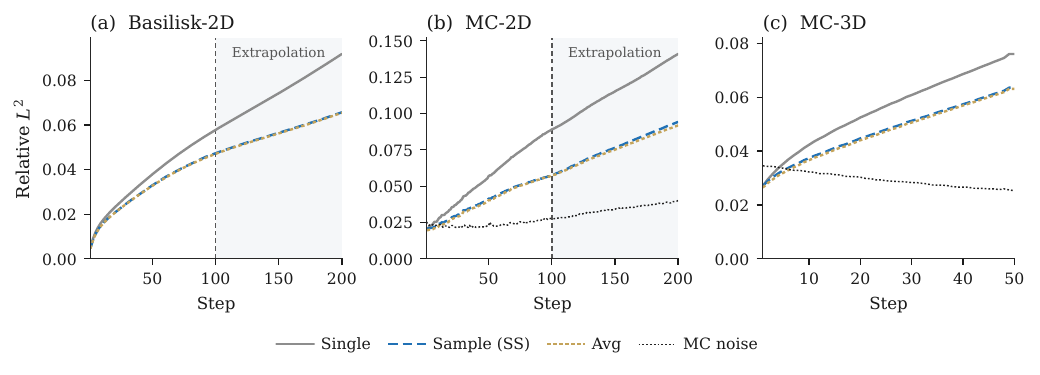}
    \caption{Relative \(L^2\) rollout error of the single-operator model with plain inference, symmetry sampling (SS), and full group averaging. The symmetry group is \(D_4\) in 2D and \(O_h\) in 3D. Black dotted curves show MC noise; shaded regions mark extrapolation beyond the training horizon.}
    \label{fig:symmetry}
\end{figure}

We first isolate the effect of inference-time symmetry on the MC-2D single-operator baseline, with model weights fixed across all variants. We compare plain rollout, stochastic \(D_4\) symmetry sampling, and full \(D_4\) group averaging. As shown in Fig.~\ref{fig:symmetry}, stochastic sampling substantially reduces rolslout error, while group averaging provides only a small additional improvement despite requiring many more forward passes per step. This supports our hypothesis that independently varying the symmetry realization can disrupt coherent error accumulation without requiring full group averaging.

We further evaluate symmetry sampling (SS) across both single-operator and split models on all benchmarks, using \(D_4\) in 2D and \(O_h\) in 3D. A symmetry element is sampled at every operator call: once per rollout step for the single model and independently for each \(\mathcal{F}_A\), \(\mathcal{F}_D\), or \(\mathcal{F}_P\) call for the split model. As summarized in Table~\ref{tab:symmetry}, SS consistently reduces rollout error across all benchmarks.

\begin{table}[t]
\centering
\caption{ Relative $L^2$ rollout error with and without symmetry sampling (SS). Percentages indicate the relative change from the corresponding model without SS. }
\label{tab:symmetry}
\small
\begin{tabular}{lcccc}
\toprule
Dataset & Single & Single+SS & Split & Split+SS \\
\midrule
Spectral-2D & $1.626\times10^{-3}$ & $\mathbf{8.424\times10^{-4}}$ ($-$48.2\%) & $2.630\times10^{-3}$ & $1.303\times10^{-3}$ ($-$50.5\%) \\
Basilisk-2D & $5.286\times10^{-2}$ & $4.176\times10^{-2}$ ($-$21.0\%) & $4.919\times10^{-2}$ & $\mathbf{3.838\times10^{-2}}$ ($-$22.0\%) \\
MC-2D & $7.78\times10^{-2}$ & $5.08\times10^{-2}$ ($-$34.7\%) & $6.95\times10^{-2}$ & $\mathbf{4.73\times10^{-2}}$ ($-$32.0\%) \\
MC-3D & $5.07\times10^{-2}$ & $4.18\times10^{-2}$ ($-$17.6\%) & $3.86\times10^{-2}$ & $\mathbf{3.33\times10^{-2}}$ ($-$13.9\%) \\
\bottomrule
\end{tabular}
\end{table}

\subsection{Corrective Cycling}

We first compare corrective cycling (CC) with unrolled and pushforward training in the MC-2D single-operator setting. CC approaches the rollout accuracy of unrolled training with substantially lower additional training time and peak GPU memory, and outperforms pushforward training in accuracy and cost under matched fine-tuning (Appendix~\ref{app:cc-cost}).

We then evaluate CC with $M=2$ operators across the benchmarks. For the single-operator model, \(\mathcal{F}_1\) and \(\mathcal{F}_2\) are alternated across successive rollout steps. For the split model, the alternation continues across the advection, diffusion, and projection stages, following $\mathcal{F}_{A1} \rightarrow \mathcal{F}_{D2} \rightarrow  \mathcal{F}_{P1} \rightarrow  \mathcal{F}_{A2} \rightarrow  \mathcal{F}_{D1} \rightarrow  \mathcal{F}_{P2} \rightarrow \cdots$. As shown in Table~\ref{tab:cc}, this strategy provides no improvement on the simpler dataset, but consistently reduces rollout error for both the single and split operators on the MC dataset with complex geometries.


\begin{table}[t]
\centering
\caption{ Relative $L^2$ rollout error with and without corrective cycling (CC). Percentages indicate the relative change from the corresponding baseline without CC. }
\label{tab:cc}
\small
\begin{tabular}{lcccc}
\toprule
Dataset & Single & Single+CC & Split & Split+CC \\
\midrule
Spectral-2D & $\mathbf{1.626\times10^{-3}}$ & $1.801\times10^{-3}$ ($+$10.8\%) & $2.630\times10^{-3}$ & $2.232\times10^{-3}$ ($-$15.1\%) \\
Basilisk-2D & $5.286\times10^{-2}$ & $5.446\times10^{-2}$ ($+$3.0\%) & $\mathbf{4.919\times10^{-2}}$ & $5.065\times10^{-2}$ ($+$3.0\%) \\
MC-2D & $7.78\times10^{-2}$ & $6.24\times10^{-2}$ ($-$19.9\%) & $6.95\times10^{-2}$ & $\mathbf{5.81\times10^{-2}}$ ($-$16.5\%) \\
MC-3D & $5.07\times10^{-2}$ & $4.00\times10^{-2}$ ($-$21.0\%) & $3.86\times10^{-2}$ & $\mathbf{3.80\times10^{-2}}$ ($-$1.7\%) \\
\bottomrule
\end{tabular}
\end{table}

\subsection{Combined Performance}

\begin{figure}[t]
    \centering
    \includegraphics[width=1\linewidth]{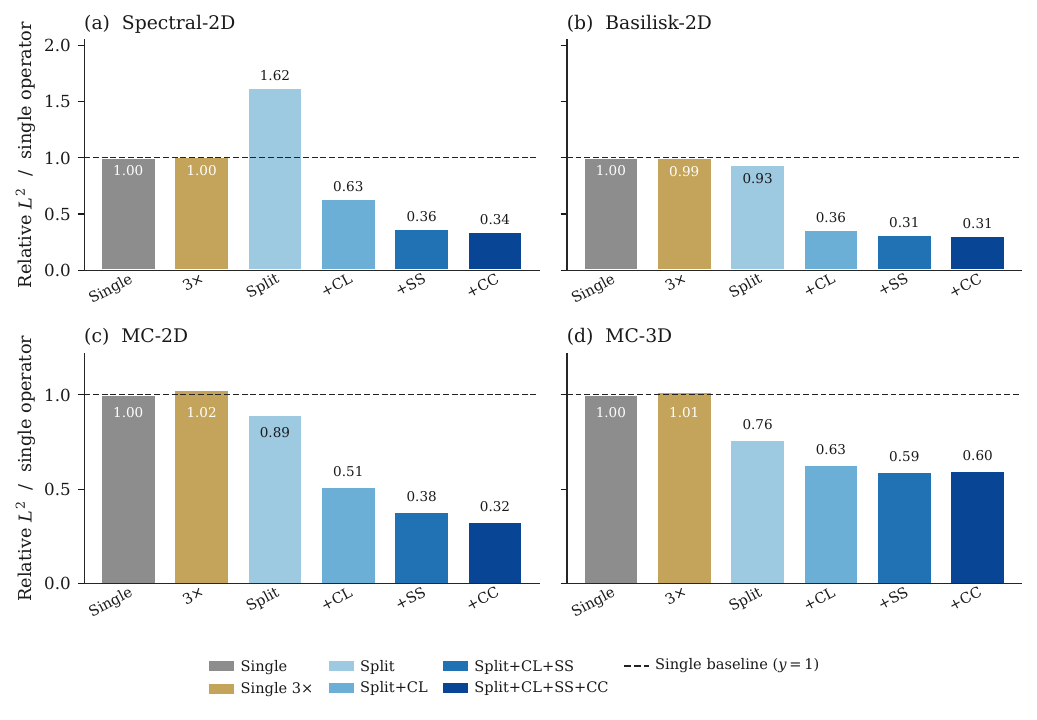}
    \caption{\textbf{Combined rollout improvements.}
    Relative \(L^2\) rollout error divided by the single-operator baseline error on each benchmark. Starting from Split, successive bars add the conditional-linear architecture (CL), symmetry sampling (SS), and corrective cycling (CC). Single \(3\times\) uses three times as many FNO layers as Single. The dashed line marks the baseline ratio of $1$.}
    \label{fig:combined}
\end{figure}

Finally, we combine the proposed components to evaluate their cumulative effect. As shown in Figure~\ref{fig:combined}, the full combination substantially reduces rollout error relative to the single-operator baseline across all benchmarks. These results show that the proposed components remain effective when used together, yielding consistent improvements in long-horizon rollout accuracy.

\section{Discussion}

\begin{figure}[t]
    \centering
    \includegraphics[width=1\linewidth]{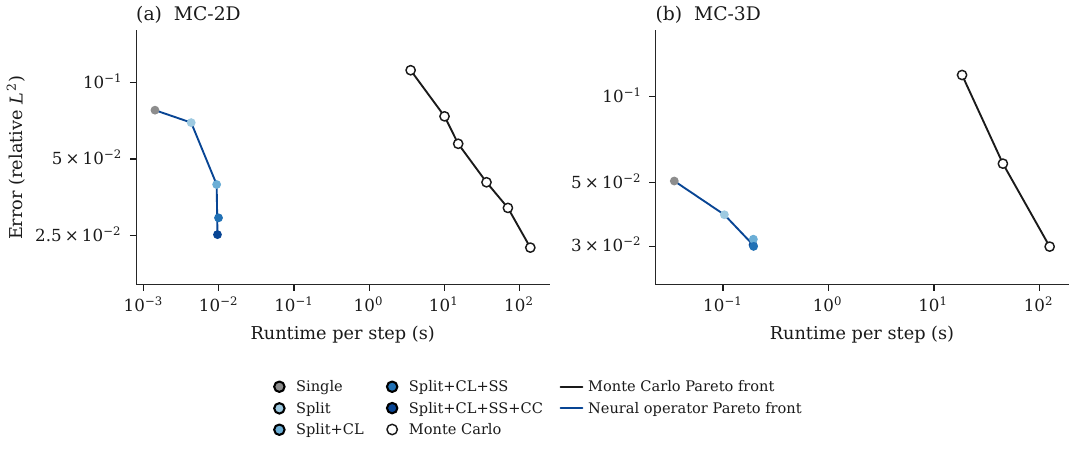}
    \caption{\textbf{Runtime--accuracy tradeoff.}
    Per-step runtime on an RTX 4090 versus relative \(L^2\) error, with logarithmic axes. Filled circles show variance-corrected neural-operator rollout errors. Open circles estimate the first-step sampling error of one MC realization from independent seeds (Appendix~\ref{app:variance-corrected-error}). MC path counts vary; other estimator settings follow Table~\ref{tab:app-mc}. Lines show Pareto fronts.}
    \label{fig:speed}
\end{figure}

Figure~\ref{fig:speed} compares the runtime--accuracy tradeoff of the learned models with VelMCFluids on MC-2D and MC-3D. MC error arises from stochastic sampling and decreases with increasing computational cost. At comparable error levels, neural-operator inference is roughly three orders of magnitude faster per step. By further lowering surrogate error, Neural Operator Splitting and the additional components make learned models viable at stricter accuracy targets, where achieving the same accuracy with MC becomes increasingly expensive.

Despite these gains, our study has several limitations. First, our experiments focus on incompressible Navier--Stokes flows with advection--diffusion--projection splitting; extending neural operator splitting to other PDE systems and fractional-step decompositions remains to be explored. The approach also relies on access to meaningful intermediate solver states for substep supervision, which may not be available in all numerical solvers. Second, the conditional-linear architecture is motivated by the structure exposed through operator splitting and is not uniformly advantageous as a standalone architecture. For nonlinear mappings such as advection and the full-step operator, we also observe reduced stability when extrapolating to Courant--Friedrichs--Lewy (CFL) numbers substantially larger than those seen during training. Its behavior under broader nonlinear and extrapolative regimes remains to be further investigated. Finally, although we evaluate across spectral, Basilisk, and Monte Carlo solvers in both 2D and 3D, broader validation across solver families and geometry distributions is left for future work.


\section{Conclusion}

We introduced Neural Operator Splitting, a framework that follows the physical substeps of a fractional-step fluid solver and learns the corresponding advection, diffusion, and pressure-projection operators from intermediate solver states. Across 2D and 3D fluid problems with complex geometries, the operator-splitting formulation consistently improves long-horizon rollout accuracy over a single full-step neural operator, while this advantage does not appear in the simpler geometry-free setting.

Building on this formulation, we further introduced a structure-matched conditional-linear architecture together with two complementary strategies for autoregressive rollout. The conditional-linear architecture matches each learned suboperator to its physical structure, incorporating the known linear structure of selected substeps while allowing the operator to adapt to geometry. Symmetry sampling varies the symmetry-transformed operator across applications to reduce coherent error accumulation. Corrective cycling trains a sequence of operators on states containing errors from preceding operators. These components provide complementary reductions in rollout error and, when combined, reduce mean rollout error by 68\% in 2D and 40\% in 3D relative to the single full-step operator (Fig~\ref{fig:combined}).

Overall, our results suggest that a full solver step need not be treated as a monolithic learning target in neural PDE solvers. Neural Operator Splitting decomposes this target into physically distinct sub-processes and provides a framework in which substep-level structural designs and rollout-level strategies can be combined. Together, these mechanisms substantially improve long-horizon prediction of fluid dynamics across complex and varying geometries. Future work could extend the evaluation to additional solver families, including OpenFOAM, explore alternative operator-splitting schemes, and extend the framework to other PDE systems.

\section*{AI use statement}

In this work, we used generative AI tools to assist with parts of method implementation and experimental work. Additionally, we used generative AI tools to polish writing and identify relevant literature. We did not use generative AI tools for other applicable tasks requiring disclosure. All AI-assisted work was reviewed and verified by the authors. We take responsibility for the final content of this work, including text, claims, and artifacts produced with the aid of generative AI.

\section*{Acknowledgements}
Anima Anandkumar and Xi Deng are both supported in part by the Bren endowed chair and the AI2050 senior fellow program at Schmidt Sciences. Hong Chul Nam is supported by a PhD fellowship for the Center for Data Science at NYU. Jianxiang Li is supported by Caltech Summer Undergraduate Research Fellowships (SURF) program.

\bibliography{iclr2027_conference}
\bibliographystyle{iclr2027_conference}

\appendix

\section{Datasets}
\label{app:data}

\begin{table}[H]
\centering
\caption{Grid resolutions, trajectory counts, and trajectory lengths (in time steps).}
\label{tab:app-datasets}
\begin{tabular}{lcccc}
\toprule
 & Spectral-2D & Basilisk-2D & MC-2D & MC-3D \\
\midrule
Grid & $64^2$ & $64^2$ & $64^2$ & $64^3$ \\
Train/validation/test & $500/50/50$ & $500/50/50$ & $2000/200/100$ & $300/30/30$ \\
Training trajectory length & $100$ & $100$ & $100$ & $50$ \\
Test trajectory length & $200$ & $200$ & $200$ & $50$ \\
\bottomrule
\end{tabular}
\end{table}

All four datasets use a domain of side length $1$, time step $\Delta t=0.02$, and kinematic viscosity $\nu=10^{-3}$. Spectral and Basilisk are doubly periodic, whereas the MC simulations use an unbounded exterior domain.

For datasets with solids, the models receive a static signed distance function (SDF) on the regular grid, negative in the solid and positive in the fluid. Basilisk evaluates its analytic solid representation at cell centers. For 2D MC, this distance is computed directly to the polygonal boundary; for 3D MC, to the normalized triangular mesh used by the solver, both at cell centers without interpolation. The velocity inside the solid is set to zero. Initial fields are lifted to approximately satisfy $\mathbf{u}\cdot\mathbf{n}=0$ at the boundary, using a harmonic stream-function lift for Basilisk and 2D MC and a potential-band lift for 3D MC. For MC, the solver applies one projection before the advection--diffusion--projection sequence.

\paragraph{Spectral-2D.}
There is no solid. The initial velocity is the curl of a stream function solved from periodic Gaussian vorticity, with vortices placed uniformly, as dipoles, or in clusters, and is scaled to a peak speed of $0.5$.

\paragraph{Basilisk-2D.}
Each trajectory has one stationary solid: a circle in training and validation, and an axis-aligned square at test. The embedded boundary is no-slip. The initial velocity is the curl of a stream function combining Fourier modes, compact vortices, and an optional uniform stream. The fluid RMS speed is drawn uniformly from $[0.10,0.30]$ and capped at $0.45$, and the projection tolerance is $10^{-3}$.

\paragraph{MC-2D.}
Each scene has one stationary solid. Training and validation geometries are analytic closed curves---circles, polygons, stars, and radial shapes---placed by a random affine transform. The test geometries are NACA airfoils outside those families. The initial velocity is the curl of a stream function combining a uniform stream, Fourier modes, vortices, shear, and wakes. The target fluid RMS speed is drawn uniformly from $[0.75,1.05]$, with speed capped at $2$.

\paragraph{MC-3D.}
Each scene has one ABC part, scaled to that its longest bounding-box side is $0.6$. The initial velocity is the curl of a vector potential assembled from a few atomic fields. The fluid RMS speed is $0.25$ and the speed is capped at $0.5$.

\begin{table}[H]
\centering
\caption{MC estimator settings. Separate training and test values are shown where they differ.}
\label{tab:app-mc}
\begin{tabular}{lcc}
\toprule
 & MC-2D & MC-3D \\
\midrule
Path length & $4$ & $4$ \\
MC paths (train/test) & $6.25\times10^{4}$ & $10^{5}/5\times10^{6}$ \\
Direct volume / pseudo-boundary samples & $1.25\times10^{5}$ & $10^{5}$ \\
Indirect volume samples & $10$ & $100$ \\
Indirect pseudo-boundary samples & $10$ & $10$ \\
Winding samples & $1$ & $0$ \\
Initial winding samples & $1024$ & $1024$ \\
\bottomrule
\end{tabular}
\end{table}

Each MC test scene has a paired reference generated with the same geometry, initial condition, and numerical parameters but an independent MC seed. These pairs estimate sampling noise as described in Appendix~\ref{app:variance-corrected-error}.


\section{Training Setup}
\label{app:setup}

\subsection{Training Loss}
For each learned operator, let \(\widehat{\mathbf u}\) and \(\mathbf u\) denote the predicted and target velocity fields, respectively. Let \(\Omega\) contain the fluid grid points for geometry-dependent data and the full grid for Spectral-2D, and let \(\mathbf x\) denote a grid point. Define the discrete norm
\begin{equation}
\|\mathbf u\|_{\ell^2(\Omega)}
:=
\left(
\sum_{\mathbf x\in\Omega}
\|\mathbf u(\mathbf x)\|_2^2
\right)^{1/2},
\end{equation}
where \(\|\cdot\|_2\) denotes the Euclidean norm over the velocity components. With \(\varepsilon=10^{-8}\), we use the relative discrete \(\ell^2\) loss
\begin{equation}
\mathcal{L}
=
\frac{
\left\|\widehat{\mathbf u}-\mathbf u\right\|_{\ell^2(\Omega)}
}{
\left\|\mathbf u\right\|_{\ell^2(\Omega)}
+\varepsilon
}.
\end{equation}
The minibatch loss is the mean of \(\mathcal{L}\) over the minibatch.

\subsection{Input Preprocessing}
\paragraph{Normalization.}
For the velocity field, we use an isotropic normalization that commutes with the symmetry transformations used for data augmentation. Let \(d\in\{2,3\}\) be the spatial dimension, \(u_c\) the velocity component indexed by \(c=1,\ldots,d\), and \(\mu_c\) and \(\sigma_c\) its training-set mean and standard deviation. With \(\varepsilon_{\mathrm{norm}}=10^{-5}\), define the shared scale \(\sigma_{\mathrm{iso}}\) and normalized component \(\widetilde{u}_c\) by
\begin{equation}
\sigma_{\mathrm{iso}}
=
\left[
\frac{1}{d}\sum_{c=1}^{d}
\left(\mu_c^2+\sigma_c^2\right)
\right]^{1/2},
\qquad
\widetilde{u}_c
=
\frac{u_c}{\sigma_{\mathrm{iso}}+\varepsilon_{\mathrm{norm}}}.
\end{equation}

Let \(s\) denote the SDF geometry channel and \(\mu_s\) and \(\sigma_s\) its training-set mean and standard deviation. Its normalized value \(\widetilde{s}\) is defined separately by
\begin{equation}
\widetilde{s}
=
\frac{s-\mu_s}{\sigma_s+\varepsilon_{\mathrm{norm}}}.
\end{equation}

\paragraph{Domain padding.}
Spectral-2D uses no domain padding, Basilisk-2D uses zero-valued padding with a padding ratio of $0.125$, and MC-2D/3D use replicate padding with the same ratio. For the MC datasets, replicate padding matches VelMCFluids' clamp-to-edge treatment of velocity queries beyond the finite computational grid used to represent the unbounded physical domain.

\subsection{Data Augmentation}
\label{app:argu}
\paragraph{Symmetry augmentation.}
During training, we independently sample a random element of \(D_4\) in 2D or \(O_h\) in 3D for each learned operator and input--target pair, and apply it to the velocity fields, SDF, and coordinates, treating velocity as a vector field. All models use this augmentation online.

\paragraph{Combination with corrective cycling.}
When generating corrective-cycling inputs offline, each preceding operator is applied with independent symmetry sampling. The resulting input--target pairs receive the same online augmentation during training.

\subsection{Conditional-linear implementation}
The conditioning branch retains the lifted feature and the output of each FNO layer. For each state-branch layer $\ell$, these features are mapped by layer-specific bias-free $1\times1$ convolutions and summed. A two-layer pointwise GELU MLP with expansion ratio $2$ maps this sum to the gate-head outputs $\boldsymbol{\delta}_\ell^{\mathrm{in}}$ and $\boldsymbol{\delta}_\ell^{\mathrm{out}}$. Each $\eta_\ell$ is learned and initialized to $0.2$. The gates are
\begin{equation}
\Gamma_\ell^{\mathrm{in}}(\mathbf{c})
=
\exp\!\left[
\eta_\ell\tanh\!\left(\boldsymbol{\delta}_\ell^{\mathrm{in}}\right)
\right],
\qquad
\Gamma_\ell^{\mathrm{out}}(\mathbf{c})
=
\exp\!\left[
\eta_\ell\tanh\!\left(\boldsymbol{\delta}_\ell^{\mathrm{out}}\right)
\right].
\end{equation}
The final gate-head layer is initialized with zero-mean weights with standard deviation $10^{-3}$ and zero bias.

The state branch uses bias-free pointwise lifting and projection, bias-free spectral convolutions, and linear skip connections, without activation.

\subsection{Model Configuration and Parameter Counts}
Table~\ref{tab:app-architecture} summarizes the architecture settings that differ across the standard FNO, conditional-linear state branch, and conditioner in the primary MC experiments. All use four Fourier layers and a projection ratio of $2$. MC inputs comprise velocity, SDF, and cell coordinates, and every learned operator predicts a residual velocity update.

\begin{table}[H]
\centering
\caption{Architecture settings for the standard FNO and the two conditional-linear branches in the MC experiments. Fourier mode counts are per spatial dimension.}
\label{tab:app-architecture}
\begin{tabular}{lccc}
\toprule
 & Standard FNO & Linear state branch & Conditioning branch \\
\midrule
Hidden width & 64 & 48 & 48 \\
Fourier modes & 24 & 24 & 20 \\
Nonlinearity & GELU & None & GELU \\
Channel MLP & Yes, expansion 0.5 & No & Yes, expansion 0.5 \\
Stabilizer & None & None & $\tanh$ \\
\bottomrule
\end{tabular}
\end{table}

Table~\ref{tab:app-parameters} reports trainable parameters per operator. The conditional-linear models have comparable or fewer parameters than the standard FNO, and their accuracy gains therefore cannot be attributed to a larger parameter count.

\begin{table}[H]
\centering
\small
\caption{Trainable parameters per learned operator. Counts include both state and conditioning branches for conditional-linear models. Single $3\times$ is the 12-layer standard FNO from Table~\ref{tab:splitting}.}
\label{tab:app-parameters}
\begin{tabular}{lcccc}
\toprule
Dataset & Standard FNO & Single $3\times$ & CL single/advection & CL diffusion/projection \\
\midrule
Spectral-2D & $5.163$M & $15.454$M & $5.048$M & $5.145$M \\
Basilisk-2D & $5.163$M & $15.454$M & $5.048$M & $5.048$M \\
MC-2D & $5.163$M & $15.454$M & $5.048$M & $5.048$M \\
MC-3D & $122.735$M & $368.169$M & $109.705$M & $109.705$M \\
\bottomrule
\end{tabular}
\end{table}

\subsection{Optimization}

We train with AdamW using a learning rate of $5\times10^{-4}$, weight decay $10^{-3}$, and batch sizes of 100 in 2D and 20 in 3D. ReduceLROnPlateau uses a factor of $0.5$, patience 20, and an absolute threshold of $10^{-5}$. Training runs for at most 2000 epochs; after epoch 50, early stopping uses patience 50 and a minimum improvement of $10^{-5}$. Compared variants use matched random seeds, and evaluation uses the checkpoint with the lowest validation loss.

\section{Variance-Corrected Error and MC Noise}
\label{app:variance-corrected-error}

For each MC test sample \(i\), representing a trajectory--time-step pair, let \(\widehat{\mathbf u}_i\) be the model prediction and let \(\mathbf u^{(1)}_i\) and \(\mathbf u^{(2)}_i\) be references generated with the same geometry, initial condition, and numerical parameters but independent MC seeds. Let \(\Omega_i\) be the fluid grid points of that sample. Using the discrete norm defined in Appendix~\ref{app:setup}, define the variance-corrected squared discrepancy \(q_i\) and squared signal \(s_i\) by
\begin{align}
q_i &=
\frac{1}{2}\left(
\left\|\widehat{\mathbf u}_i-\mathbf u^{(1)}_i\right\|_{\ell^2(\Omega_i)}^2
+
\left\|\widehat{\mathbf u}_i-\mathbf u^{(2)}_i\right\|_{\ell^2(\Omega_i)}^2
-
\left\|\mathbf u^{(1)}_i-\mathbf u^{(2)}_i\right\|_{\ell^2(\Omega_i)}^2
\right),\\
s_i &=
\frac{1}{2}\left(
\left\|\mathbf u^{(1)}_i\right\|_{\ell^2(\Omega_i)}^2
+
\left\|\mathbf u^{(2)}_i\right\|_{\ell^2(\Omega_i)}^2
-
\left\|\mathbf u^{(1)}_i-\mathbf u^{(2)}_i\right\|_{\ell^2(\Omega_i)}^2
\right).
\end{align}

Let \(\mathbf u_i\) be the common expectation of the two MC references. If their sampling errors are mutually uncorrelated and uncorrelated with the model error, then
\begin{equation}
\mathbb E[q_i]
=
\left\|\widehat{\mathbf u}_i-\mathbf u_i\right\|_{\ell^2(\Omega_i)}^2,
\qquad
\mathbb E[s_i]
=
\left\|\mathbf u_i\right\|_{\ell^2(\Omega_i)}^2.
\end{equation}
The correction removes the contribution of MC sampling variance in expectation but not bias shared by the two references.

Let \(N\) be the number of test samples in the evaluation window and \(\Delta V_i\) the grid-cell area in 2D or volume in 3D for sample \(i\). We report the pooled absolute and relative errors \(E_{\mathrm{abs}}\) and \(E_{\mathrm{rel}}\), respectively:
\begin{equation}
E_{\mathrm{abs}}=
\sqrt{\max\left(\frac{1}{N}\sum_i\Delta V_i q_i,0\right)},
\qquad
E_{\mathrm{rel}}=
\sqrt{\frac{\max\left(\sum_i q_i,0\right)}{\sum_i s_i}}.
\end{equation}
The sums include all \(N\) records with equal weight, and the relative error is reported only when \(\sum_i s_i>0\). Within each benchmark, all records share the same grid-cell area or volume, which cancels in the relative error. Signed values of \(q_i\) and \(s_i\) are retained during aggregation; the nonnegative clamp is applied to the pooled squared error before taking its square root.

\paragraph{Standard rollout error and MC noise.}
For each time step in Figs.~\ref{fig:cl} and~\ref{fig:symmetry}, we pool the squared norms over the test trajectories. With sums restricted to records at that step, the standard relative error is
\begin{equation}
\sqrt{\frac{\sum_i\left\|\widehat{\mathbf u}_i-\mathbf u^{(1)}_i\right\|_{\ell^2(\Omega_i)}^2}
{\sum_i\left\|\mathbf u^{(1)}_i\right\|_{\ell^2(\Omega_i)}^2}},
\end{equation}
and MC noise is
\begin{equation}
\sqrt{\frac{\sum_i\left\|\mathbf u^{(1)}_i-\mathbf u^{(2)}_i\right\|_{\ell^2(\Omega_i)}^2}
{\sum_i\left\|\mathbf u^{(1)}_i\right\|_{\ell^2(\Omega_i)}^2}}.
\end{equation}
This discrepancy includes the sampling variance of both references. The displayed curves use a three-point temporal median filter, with endpoints unchanged.

For the open circles in Fig.~\ref{fig:speed}, we use first-step records to estimate the relative sampling error of one MC realization:
\begin{equation}
\sqrt{\frac{\sum_i\left\|\mathbf u^{(1)}_i-\mathbf u^{(2)}_i\right\|_{\ell^2(\Omega_i)}^2}
{2\sum_i s_i}},
\qquad \sum_i s_i>0.
\end{equation}
The factor of \(1/2\) converts the squared difference of independent references with equal sampling budgets to the sampling variance of one realization; the denominator uses the variance-corrected signal defined above.

\section{Additional Evaluation Metrics}
\label{app:additional-metrics}

Tables~\ref{tab:app-metrics-split}--\ref{tab:app-metrics-combined} give a counterpart to each results table in the main text. The Split and CL tables additionally report one-step relative $L^2$ prediction error, evaluated from ground-truth input states. All five tables report rollout mean squared error (MSE) and absolute divergence over the same test trajectories and rollout horizons as their main-text counterparts.

We retain the sample index \(i\), sample count \(N\), fluid grid \(\Omega_i\), and cell area or volume \(\Delta V_i\) from Appendix~\ref{app:variance-corrected-error}. For deterministic benchmarks, \(\mathbf u_i\) denotes the deterministic reference field, and rollout MSE is the average physical squared velocity error:
\begin{equation}
\mathrm{MSE}
=
\frac{1}{N}\sum_i
\Delta V_i
\left\|
\widehat{\mathbf u}_i-\mathbf u_i
\right\|_{\ell^2(\Omega_i)}^2.
\end{equation}
For MC benchmarks, we replace the squared discrepancy with the paired-reference corrected quantity \(q_i\) from Appendix~\ref{app:variance-corrected-error}, and apply the nonnegative clamp after averaging:
\begin{equation}
\mathrm{MSE}
=
\max\left(
\frac{1}{N}\sum_i \Delta V_i q_i,
0
\right)
=
E_{\mathrm{abs}}^2.
\end{equation}
Thus, MSE has units of squared velocity integrated over the unit domain.

Absolute divergence is
\begin{equation}
\left(
\frac{1}{N}\sum_i
\left\|
\nabla\cdot\widehat{\mathbf u}_i
\right\|_{L^2(\Omega_i)}^2
\right)^{1/2},
\end{equation}
where the norm is evaluated on the common valid stencil support within \(\Omega_i\). This measures the prediction's own divergence, rather than a divergence error against the reference, and no MC correction is applied.

The one-step relative \(L^2\) prediction error follows the pooled convention of Appendix~\ref{app:variance-corrected-error} for MC data and the corresponding pooled error-to-signal ratio for deterministic data.

\begin{table}[H]
\centering
\setlength{\tabcolsep}{3pt}
\small
\caption{One-step relative $L^2$ error, rollout MSE, and absolute divergence for the models in Table~\ref{tab:splitting}.}
\label{tab:app-metrics-split}
\begin{tabular}{llccc}
\toprule
Dataset & Metric & Single & Single $3\times$ & Split \\ 
\midrule
Spectral-2D & One-step rel. $L^2$ & $5.206\times10^{-5}$ & $\mathbf{4.944\times10^{-5}}$ & $7.723\times10^{-5}$ \\ 
 & Rollout MSE & $\mathbf{1.005\times10^{-7}}$ & $1.015\times10^{-7}$ & $2.631\times10^{-7}$ \\ 
 & Rollout abs div & $5.029\times10^{-3}$ & $\mathbf{3.663\times10^{-3}}$ & $5.157\times10^{-3}$ \\ 
\midrule
Basilisk-2D & One-step rel. $L^2$ & $1.533\times10^{-3}$ & $\mathbf{1.499\times10^{-3}}$ & $2.138\times10^{-3}$ \\ 
 & Rollout MSE & $3.099\times10^{-5}$ & $3.058\times10^{-5}$ & $\mathbf{2.683\times10^{-5}}$ \\ 
 & Rollout abs div & $5.826\times10^{-2}$ & $\mathbf{5.605\times10^{-2}}$ & $6.602\times10^{-2}$ \\ 
\midrule
MC-2D & One-step rel. $L^2$ & $4.132\times10^{-3}$ & $\mathbf{4.068\times10^{-3}}$ & $4.137\times10^{-3}$ \\ 
 & Rollout MSE & $1.773\times10^{-3}$ & $1.852\times10^{-3}$ & $\mathbf{1.415\times10^{-3}}$ \\ 
 & Rollout abs div & $1.146\times10^{-1}$ & $\mathbf{1.128\times10^{-1}}$ & $1.153\times10^{-1}$ \\ 
\midrule
MC-3D & One-step rel. $L^2$ & $6.058\times10^{-3}$ & $5.602\times10^{-3}$ & $\mathbf{5.828\times10^{-3}}$ \\ 
 & Rollout MSE & $2.623\times10^{-5}$ & $2.694\times10^{-5}$ & $\mathbf{1.525\times10^{-5}}$ \\ 
 & Rollout abs div & $2.521\times10^{-2}$ & $2.792\times10^{-2}$ & $\mathbf{2.288\times10^{-2}}$ \\ 
\bottomrule
\end{tabular}
\end{table}

\begin{table}[H]
\centering
\setlength{\tabcolsep}{3pt}
\small
\caption{One-step relative $L^2$ error, rollout MSE, and absolute divergence for the models in Table~\ref{tab:cl}.}
\label{tab:app-metrics-cl}
\begin{tabular}{llcccc}
\toprule
Dataset & Metric & Single & CL Single & Split & CL Split \\ 
\midrule
Spectral-2D & One-step rel. $L^2$ & $5.206\times10^{-5}$ & $\mathbf{4.984\times10^{-5}}$ & $7.723\times10^{-5}$ & $6.561\times10^{-5}$ \\ 
 & Rollout MSE & $1.005\times10^{-7}$ & $2.220\times10^{-7}$ & $2.631\times10^{-7}$ & $\mathbf{4.045\times10^{-8}}$ \\ 
 & Rollout abs div & $5.029\times10^{-3}$ & $5.519\times10^{-3}$ & $5.157\times10^{-3}$ & $\mathbf{4.276\times10^{-3}}$ \\ 
\midrule
Basilisk-2D & One-step rel. $L^2$ & $1.533\times10^{-3}$ & $\mathbf{1.151\times10^{-3}}$ & $2.138\times10^{-3}$ & $2.305\times10^{-3}$ \\ 
 & Rollout MSE & $3.099\times10^{-5}$ & $2.772\times10^{-5}$ & $2.683\times10^{-5}$ & $\mathbf{4.010\times10^{-6}}$ \\ 
 & Rollout abs div & $5.826\times10^{-2}$ & $5.555\times10^{-2}$ & $6.602\times10^{-2}$ & $\mathbf{4.160\times10^{-2}}$ \\ 
\midrule
MC-2D & One-step rel. $L^2$ & $4.132\times10^{-3}$ & $3.103\times10^{-3}$ & $4.137\times10^{-3}$ & $\mathbf{2.661\times10^{-3}}$ \\ 
 & Rollout MSE & $1.773\times10^{-3}$ & $1.687\times10^{-3}$ & $1.415\times10^{-3}$ & $\mathbf{4.611\times10^{-4}}$ \\ 
 & Rollout abs div & $1.146\times10^{-1}$ & $9.863\times10^{-2}$ & $1.153\times10^{-1}$ & $\mathbf{9.836\times10^{-2}}$ \\ 
\midrule
MC-3D & One-step rel. $L^2$ & $6.058\times10^{-3}$ & $5.362\times10^{-3}$ & $5.828\times10^{-3}$ & $\mathbf{5.122\times10^{-3}}$ \\ 
 & Rollout MSE & $2.623\times10^{-5}$ & $1.709\times10^{-5}$ & $1.525\times10^{-5}$ & $\mathbf{1.028\times10^{-5}}$ \\ 
 & Rollout abs div & $2.521\times10^{-2}$ & $1.998\times10^{-2}$ & $2.288\times10^{-2}$ & $\mathbf{1.760\times10^{-2}}$ \\ 
\bottomrule
\end{tabular}
\end{table}

\begin{table}[H]
\centering
\setlength{\tabcolsep}{3pt}
\small
\caption{Rollout MSE and absolute divergence with and without symmetry sampling, corresponding to Table~\ref{tab:symmetry}.}
\label{tab:app-metrics-ss}
\begin{tabular}{llcccc}
\toprule
Dataset & Metric & Single & Single+SS & Split & Split+SS \\ 
\midrule
Spectral-2D & Rollout MSE & $1.005\times10^{-7}$ & $\mathbf{2.699\times10^{-8}}$ & $2.631\times10^{-7}$ & $6.456\times10^{-8}$ \\ 
 & Rollout abs div & $5.029\times10^{-3}$ & $\mathbf{3.954\times10^{-3}}$ & $5.157\times10^{-3}$ & $3.998\times10^{-3}$ \\ 
\midrule
Basilisk-2D & Rollout MSE & $3.099\times10^{-5}$ & $1.934\times10^{-5}$ & $2.683\times10^{-5}$ & $\mathbf{1.634\times10^{-5}}$ \\ 
 & Rollout abs div & $5.826\times10^{-2}$ & $\mathbf{3.145\times10^{-2}}$ & $6.602\times10^{-2}$ & $3.952\times10^{-2}$ \\ 
\midrule
MC-2D & Rollout MSE & $1.773\times10^{-3}$ & $7.566\times10^{-4}$ & $1.415\times10^{-3}$ & $\mathbf{6.547\times10^{-4}}$ \\ 
 & Rollout abs div & $1.146\times10^{-1}$ & $\mathbf{1.082\times10^{-1}}$ & $1.153\times10^{-1}$ & $1.119\times10^{-1}$ \\ 
\midrule
MC-3D & Rollout MSE & $2.623\times10^{-5}$ & $1.783\times10^{-5}$ & $1.525\times10^{-5}$ & $\mathbf{1.130\times10^{-5}}$ \\ 
 & Rollout abs div & $2.521\times10^{-2}$ & $1.925\times10^{-2}$ & $2.288\times10^{-2}$ & $\mathbf{1.922\times10^{-2}}$ \\ 
\bottomrule
\end{tabular}
\end{table}

\begin{table}[H]
\centering
\setlength{\tabcolsep}{3pt}
\small
\caption{Rollout MSE and absolute divergence with and without corrective cycling, corresponding to Table~\ref{tab:cc}.}
\label{tab:app-metrics-cc}
\begin{tabular}{llcccc}
\toprule
Dataset & Metric & Single & Single+CC & Split & Split+CC \\ 
\midrule
Spectral-2D & Rollout MSE & $\mathbf{1.005\times10^{-7}}$ & $1.233\times10^{-7}$ & $2.631\times10^{-7}$ & $1.894\times10^{-7}$ \\ 
 & Rollout abs div & $5.029\times10^{-3}$ & $5.286\times10^{-3}$ & $5.157\times10^{-3}$ & $\mathbf{4.886\times10^{-3}}$ \\ 
\midrule
Basilisk-2D & Rollout MSE & $3.099\times10^{-5}$ & $3.289\times10^{-5}$ & $\mathbf{2.683\times10^{-5}}$ & $2.846\times10^{-5}$ \\ 
 & Rollout abs div & $5.826\times10^{-2}$ & $\mathbf{5.335\times10^{-2}}$ & $6.602\times10^{-2}$ & $6.417\times10^{-2}$ \\ 
\midrule
MC-2D & Rollout MSE & $1.773\times10^{-3}$ & $1.139\times10^{-3}$ & $1.415\times10^{-3}$ & $\mathbf{9.874\times10^{-4}}$ \\ 
 & Rollout abs div & $1.146\times10^{-1}$ & $\mathbf{1.105\times10^{-1}}$ & $1.153\times10^{-1}$ & $1.138\times10^{-1}$ \\ 
\midrule
MC-3D & Rollout MSE & $2.623\times10^{-5}$ & $1.637\times10^{-5}$ & $1.525\times10^{-5}$ & $\mathbf{1.473\times10^{-5}}$ \\ 
 & Rollout abs div & $2.521\times10^{-2}$ & $\mathbf{2.027\times10^{-2}}$ & $2.288\times10^{-2}$ & $2.084\times10^{-2}$ \\ 
\bottomrule
\end{tabular}
\end{table}

\begin{table}[H]
\centering
\setlength{\tabcolsep}{3pt}
\small
\caption{Relative $L^2$ rollout error, MSE, and absolute divergence for the model configurations in Fig.~\ref{fig:combined}.}
\label{tab:app-metrics-combined}
\begin{tabular}{llccccc}
\toprule
Dataset & Metric & Single & Split & CL Split & \makecell{Split+CL\\+SS} & \makecell{Split+CL\\+SS+CC} \\ 
\midrule
Spectral-2D & Rollout rel. $L^2$ & $1.626\times10^{-3}$ & $2.630\times10^{-3}$ & $1.031\times10^{-3}$ & $5.908\times10^{-4}$ & $\mathbf{5.511\times10^{-4}}$ \\ 
 & Rollout MSE & $1.005\times10^{-7}$ & $2.631\times10^{-7}$ & $4.045\times10^{-8}$ & $1.327\times10^{-8}$ & $\mathbf{1.155\times10^{-8}}$ \\ 
 & Rollout abs div & $5.029\times10^{-3}$ & $5.157\times10^{-3}$ & $4.276\times10^{-3}$ & $4.108\times10^{-3}$ & $\mathbf{3.629\times10^{-3}}$ \\ 
\midrule
Basilisk-2D & Rollout rel. $L^2$ & $5.286\times10^{-2}$ & $4.919\times10^{-2}$ & $1.901\times10^{-2}$ & $1.660\times10^{-2}$ & $\mathbf{1.616\times10^{-2}}$ \\ 
 & Rollout MSE & $3.099\times10^{-5}$ & $2.683\times10^{-5}$ & $4.010\times10^{-6}$ & $3.057\times10^{-6}$ & $\mathbf{2.897\times10^{-6}}$ \\ 
 & Rollout abs div & $5.826\times10^{-2}$ & $6.602\times10^{-2}$ & $4.160\times10^{-2}$ & $3.955\times10^{-2}$ & $\mathbf{3.719\times10^{-2}}$ \\ 
\midrule
MC-2D & Rollout rel. $L^2$ & $7.784\times10^{-2}$ & $6.953\times10^{-2}$ & $3.969\times10^{-2}$ & $2.934\times10^{-2}$ & $\mathbf{2.520\times10^{-2}}$ \\ 
 & Rollout MSE & $1.773\times10^{-3}$ & $1.415\times10^{-3}$ & $4.611\times10^{-4}$ & $2.519\times10^{-4}$ & $\mathbf{1.858\times10^{-4}}$ \\ 
 & Rollout abs div & $1.146\times10^{-1}$ & $1.153\times10^{-1}$ & $9.836\times10^{-2}$ & $9.711\times10^{-2}$ & $\mathbf{9.606\times10^{-2}}$ \\ 
\midrule
MC-3D & Rollout rel. $L^2$ & $5.069\times10^{-2}$ & $3.865\times10^{-2}$ & $3.174\times10^{-2}$ & $\mathbf{2.995\times10^{-2}}$ & $3.032\times10^{-2}$ \\ 
 & Rollout MSE & $2.623\times10^{-5}$ & $1.525\times10^{-5}$ & $1.028\times10^{-5}$ & $\mathbf{9.157\times10^{-6}}$ & $9.386\times10^{-6}$ \\ 
 & Rollout abs div & $2.521\times10^{-2}$ & $2.288\times10^{-2}$ & $1.760\times10^{-2}$ & $1.675\times10^{-2}$ & $\mathbf{1.626\times10^{-2}}$ \\ 
\bottomrule
\end{tabular}
\end{table}

\section{Adjacent Operator-Error Correlation}
\label{app:error-correlation}

To examine how symmetry sampling affects errors during rollout, we compare the Spectral-2D single FNO with and without SS, keeping the weights fixed, and also evaluate two-operator CC. Let \(\mathcal S\) denote one step of the reference solver. The local operator error at call \(k\) is
\begin{equation}
\mathbf e_k=\hat{\mathbf u}_{k+1}-\mathcal S(\hat{\mathbf u}_k).
\end{equation}
The model and reference solver thus start from the same predicted state \(\hat{\mathbf u}_k\), and the rollout continues with the model output \(\hat{\mathbf u}_{k+1}\). Unlike the one-step prediction error evaluated from ground-truth inputs in Appendix~\ref{app:additional-metrics}, this local operator error is evaluated along the predicted trajectory. We measure its magnitude and the cosine similarity of successive errors,
\begin{equation}
\operatorname{mean}\|\mathbf e_k\|,
\qquad
\operatorname{mean}
\frac{\langle\mathbf e_{k-1},\mathbf e_k\rangle}
{\|\mathbf e_{k-1}\|\|\mathbf e_k\|},
\end{equation}
where the averages run over test trajectories and operator calls. Norms and inner products are Euclidean over all grid points and velocity components. The cosine measures how consistently successive calls produce errors in the same spatial direction.

Table~\ref{tab:app-correlation} reports results on 50 test trajectories over 100 steps, with SS sampling independently at each call. SS reduces error correlation and mean local relative error, the latter possibly reflecting less rollout distribution shift. CC lowers correlation but increases local error, potentially because its second operator is harder to optimize on predicted inputs; this trade-off yields worse rollout accuracy here.

\begin{table}[H]
\centering
\caption{Local operator errors and rollout error on 50 Spectral-2D test trajectories over 100 steps. Mean local error averages 100 operator calls per trajectory; mean cosine similarity averages 99 consecutive error pairs. Rollout error is the pooled relative \(L^2\) error over these trajectories and steps.}
\label{tab:app-correlation}
\begin{tabular}{lccc}
\toprule
Model & Mean local operator error & Mean cosine & Rollout error \\
\midrule
Single & $5.03\times10^{-3}$ & 0.997 & $1.17\times10^{-3}$ \\
Single+SS & $2.92\times10^{-3}$ & 0.947 & $6.86\times10^{-4}$ \\
Single+CC & $5.63\times10^{-3}$ & 0.990 & $1.27\times10^{-3}$ \\
\bottomrule
\end{tabular}
\end{table}

\begin{figure}[t]
\centering
\includegraphics[width=\linewidth]{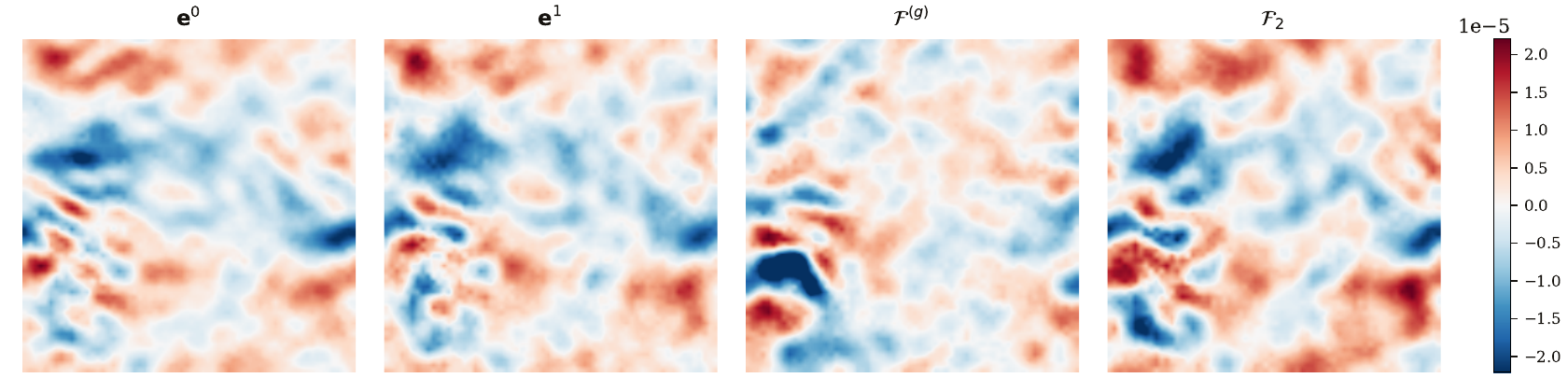}
\caption{The $x$-component of two successive local operator errors, with a shared color scale. All variants share the first call, which uses no symmetry transform. The second call uses the baseline, SS, or CC. For this illustration, SS uses the $180^\circ$ rotation that minimizes the absolute cosine similarity between consecutive error fields among non-identity \(D_4\) transforms; Table~\ref{tab:app-correlation} uses independent random symmetry samples.}
\label{fig:app-error-fields}
\end{figure}

Figure~\ref{fig:app-error-fields} shows two successive local errors on one trajectory. The baseline produces nearly identical error patterns across the two calls, while both SS and CC alter the spatial distribution of the error.

\section{Training with Monte Carlo Noise}
\label{app:mc-noise-regularization}

We generate paired MC and deterministic trajectories for a two-dimensional problem without solids in an unbounded domain. As in the other benchmarks, we use a \(64^2\) grid on the unit square, \(\Delta t=0.02\), and \(\nu=10^{-3}\). Initial fields combine background flow, Fourier modes, vortices, shear, and wakes. Both solvers clamp velocity queries outside the grid and project the initial field before the first time step. The dataset contains 500/50/50 train/validation/test initial fields, with 100 training steps and 200 test steps. Models receive velocity and coordinates, without a geometry channel.

The deterministic BIE solver follows the MC formulation, replacing stochastic integral estimates with deterministic quadrature. Both solvers use the same semi-Lagrangian RK3 advection scheme. For pressure projection, the BIE solver evaluates the volume and pseudo-boundary integrals using grid convolution and boundary quadrature. For diffusion, it convolves the heat kernel with the MC estimator's bilinear interpolation basis. Over the 200-step test trajectories, its pooled variance-corrected relative \(L^2\) error against the MC references is \(1.57\%\) (Appendix~\ref{app:variance-corrected-error}).

We train the same single-step FNO on (i) MC trajectories, (ii) BIE trajectories, and (iii) BIE trajectories with Gaussian input noise. In the third setting, we add independent Gaussian noise to each input velocity component and leave the targets unperturbed. We set the noise standard deviation to \(4.6\times10^{-3}\) in physical velocity units, matching the pooled per-component RMS MC sampling noise estimated from paired references over the first 100 steps.

All runs use $D_4$ training augmentation; the 50- and 10-trajectory subsets select matching initial fields from the 500-trajectory set. Validation and test inputs receive no added Gaussian noise. We evaluate 200-step autoregressive rollouts on both test sets, sampling a $D_4$ transform at each prediction step.

\begin{figure}[t]
\centering
\includegraphics[width=\linewidth]{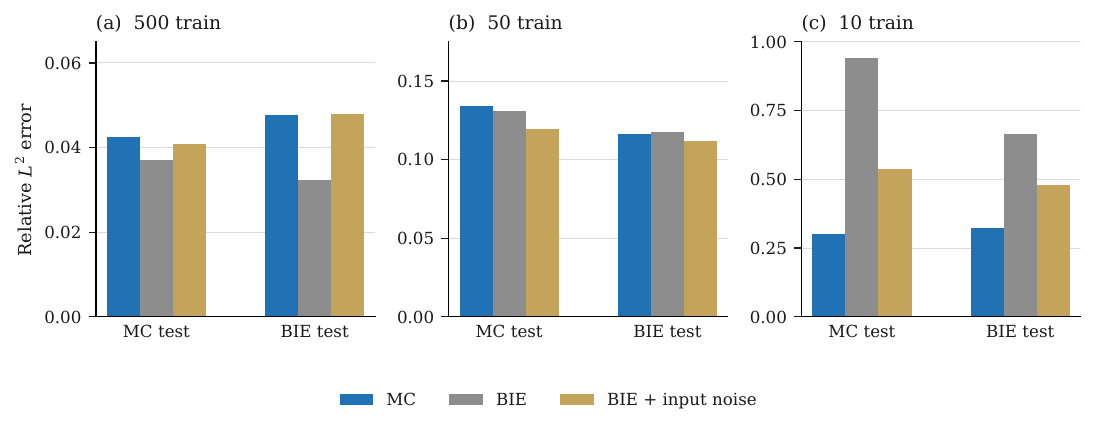}
\caption{Relative $L^2$ rollout error after training on 500, 50, or 10 trajectories. MC test errors use the paired-reference pooled variance correction of Appendix~\ref{app:variance-corrected-error}; BIE test errors use standard relative $L^2$ error.}
\label{fig:mc-noise-training}
\end{figure}

Figure~\ref{fig:mc-noise-training} shows that adding Gaussian input noise to BIE training brings its rollout errors much closer to those of MC training. MC training performs best with 10 trajectories. Gaussian input noise substantially improves BIE training with limited data but introduces a modest disadvantage with the full training set, consistent with a regularization effect. These trends are similar on the BIE and MC test sets.

\section{Corrective Cycling versus Unrolled and Pushforward Training}
\label{app:cc-cost}

We compare two-operator CC with two-step unrolled and single-step pushforward training on MC-2D. Unrolled training recursively applies the same operator for two time steps and minimizes the average loss against the corresponding ground-truth states, backpropagating through both steps. Pushforward training uses equal numbers of ground-truth inputs and inputs generated by advancing ground-truth states by one step. For both input types, it minimizes a one-step loss against the corresponding next-step ground-truth state. The predicted inputs are regenerated at the start of each epoch using a frozen model snapshot, without backpropagating through their generation. Both baselines apply a single operator throughout inference rollouts.

Full training follows the standard optimization schedule of Appendix~\ref{app:setup}, with a separately trained base operator for CC. Fine-tuning initializes the trainable operator from the same pretrained base and runs for 100 epochs at one-tenth of the original learning rate, with cosine decay and no early stopping. Table~\ref{tab:app-cc-cost} reports 200-step rollout error and training cost. All runs use $D_4$ symmetry augmentation during training and are evaluated without symmetry sampling (SS). Errors use the pooled variance-corrected metric of Appendix~\ref{app:variance-corrected-error}.

\begin{table}[H]
\centering
\setlength{\tabcolsep}{3pt}
\caption{Training cost on one NVIDIA B200 and 200-step relative $L^2$ rollout error on MC-2D NACA. Times are in hours:minutes:seconds. Fine-tuning and CC times exclude base training; CC times include data export. Full CC takes 8:59:55 including base training, still less than unrolled training. GPU memory is the peak training allocation. Bold marks the lowest values within each comparison.}
\label{tab:app-cc-cost}
\begin{tabular}{llccc}
\toprule
Comparison & Method & Time & Rollout error & Peak GPU (GB) \\
\midrule
Reference & Base & 4:14:16 & $0.078$ & $6.727$ \\
\midrule
Full training & Unrolled & 9:13:05 & $\mathbf{0.055}$ & $11.157$ \\
 & CC & \textbf{4:45:39} & $0.062$ & $\mathbf{6.727}$ \\
\midrule
Fine-tuning & Unrolled & 2:51:35 & $\mathbf{0.073}$ & $11.073$ \\
 & CC & \textbf{1:38:19} & $0.074$ & $\mathbf{6.727}$ \\
 & Pushforward & 2:18:18 & $0.076$ & $12.801$ \\
\bottomrule
\end{tabular}
\end{table}

CC approaches the rollout accuracy of unrolled training with lower training time and substantially less peak GPU memory; in fine-tuning, it also improves on pushforward in accuracy and cost. CC trains each successive operator on states generated by its frozen predecessors, allowing it to adapt to the accumulated prediction errors in its inputs.

\end{document}

%% file: math_commands.tex
\usepackage{bm, amsmath,amssymb,graphicx,tikz}
\usetikzlibrary{arrows.meta,calc,fit,positioning,shapes.geometric}

\definecolor{ink}{HTML}{27313A}
\definecolor{subink}{HTML}{66727C}
\definecolor{linegray}{HTML}{AAB5BD}
\definecolor{panelbg}{HTML}{FAFBFC}
\definecolor{plum}{HTML}{756778}
\definecolor{plumlight}{HTML}{EEEAF0}
\definecolor{steel}{HTML}{526F82}
\definecolor{steellight}{HTML}{E9EFF2}
\definecolor{teal}{HTML}{5D8584}
\definecolor{teallight}{HTML}{E7F0EF}
\definecolor{sand}{HTML}{A28E72}
\definecolor{sandlight}{HTML}{F2EFEA}

\newcommand{\figfont}{\sffamily}

\newcommand{\velocityimage}{%
  \includegraphics[width=1.23cm,height=1.12cm]{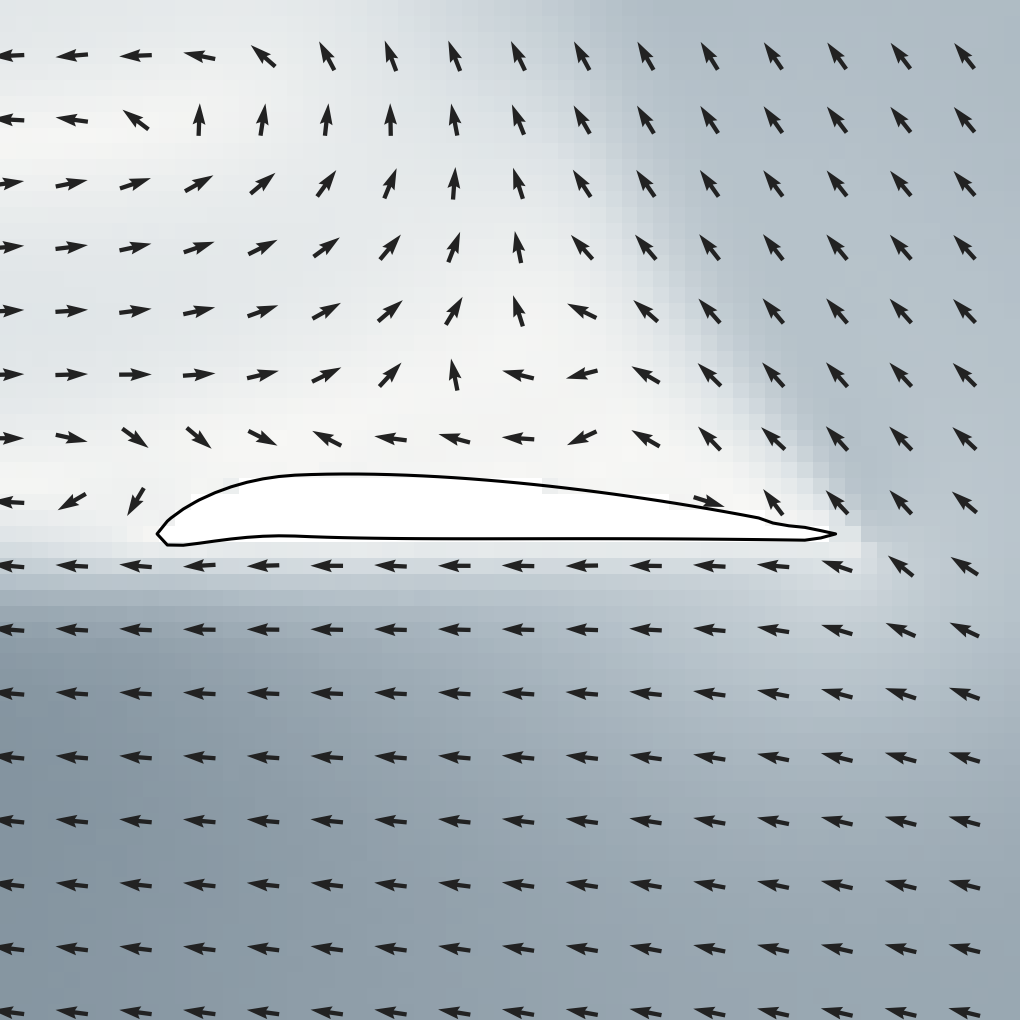}}
\newcommand{\mcadvectionimage}{%
  \includegraphics[width=1.35cm,height=1.35cm]{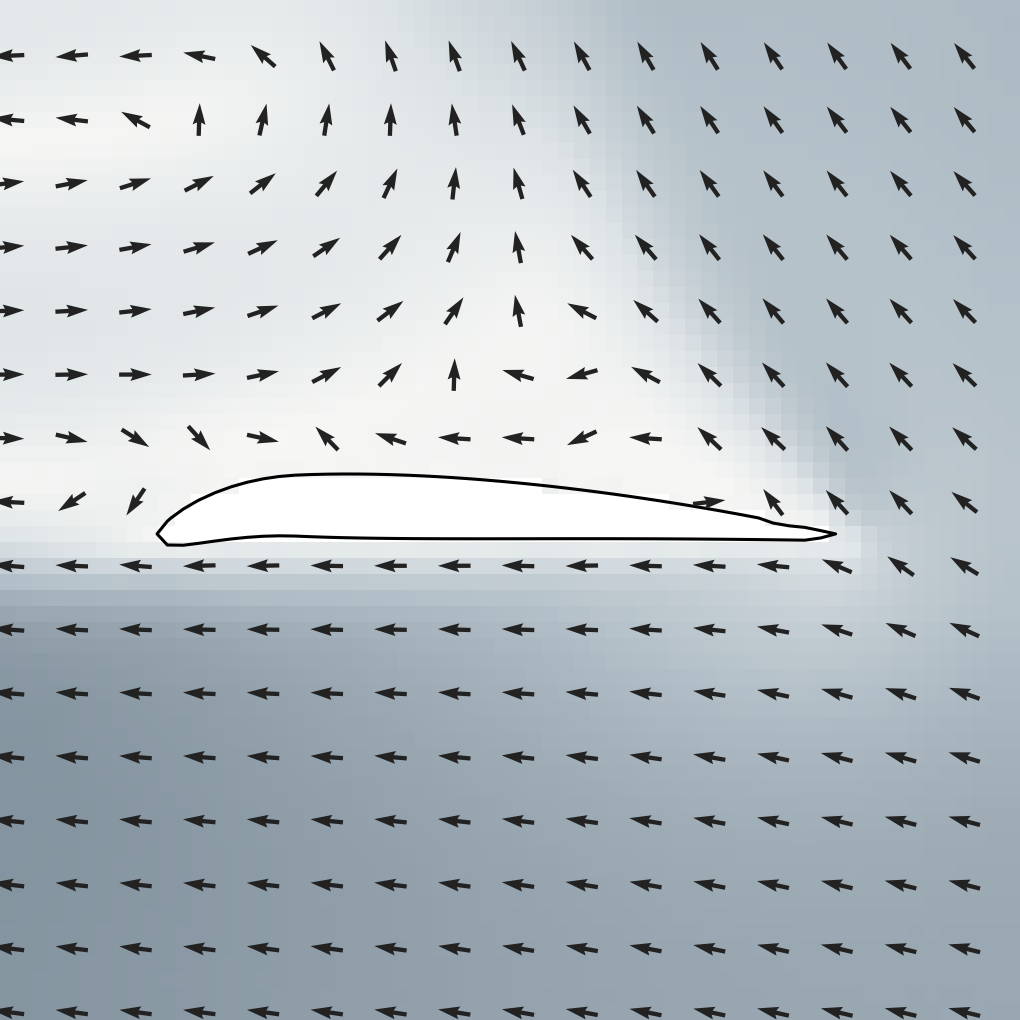}}
\newcommand{\mcdiffusionimage}{%
  \includegraphics[width=1.35cm,height=1.35cm]{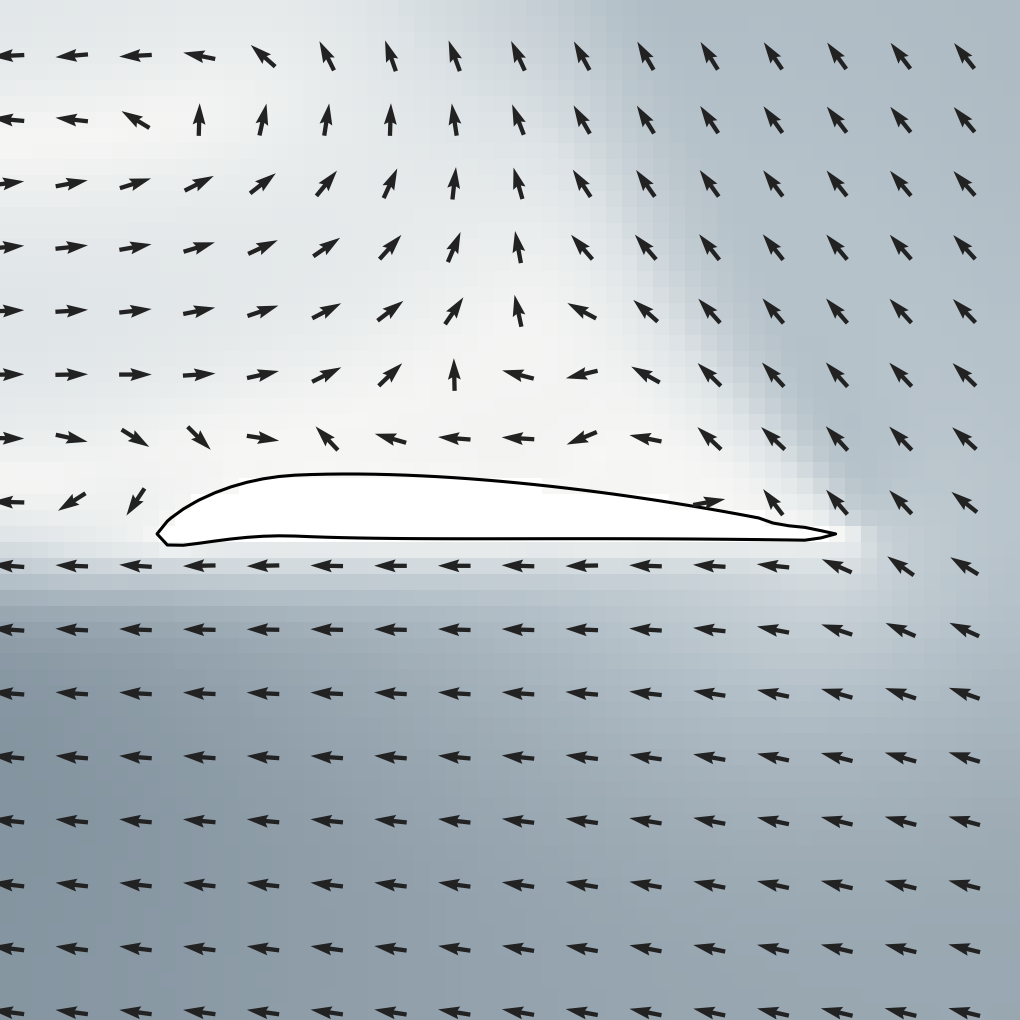}}
\newcommand{\mcprojectionimage}{%
  \includegraphics[width=1.35cm,height=1.35cm]{fig/naca_traj0_step100_velocity.png}}

\newcommand{\airfoil}[1]{%
  \begin{scope}[shift={#1}]
    \path[fill=plum]
      (0,0) .. controls (0.32,0.28) and (1.12,0.24) .. (1.72,0.02)
      .. controls (1.12,0.06) and (0.36,0.09) .. (0,0) -- cycle;
  \end{scope}%
}

\def\eqref#1{equation~\ref{#1}}

\def\1{\bm{1}}

\DeclareMathAlphabet{\mathsfit}{\encodingdefault}{\sfdefault}{m}{sl}
\SetMathAlphabet{\mathsfit}{bold}{\encodingdefault}{\sfdefault}{bx}{n}



%% file: fig/pipeline.tex
\begin{tikzpicture}[
  x=1cm,y=1cm, font=\figfont\small, text=ink,
  >={Stealth[length=2.0mm,width=1.45mm]},
  panel/.style={draw=linegray,fill=panelbg,rounded corners=1.7mm,line width=.55pt},
  title/.style={font=\figfont\bfseries\small,anchor=west},
  stage/.style={draw=plum,fill=plumlight,rounded corners=.8mm,
    minimum width=1.55cm,minimum height=.78cm,align=center,line width=.55pt},
  mcbox/.style={draw=teal,fill=teallight,rounded corners=.7mm,
    minimum width=1.70cm,minimum height=.60cm,align=center,line width=.55pt},
  softbox/.style={draw=plum,fill=white,rounded corners=.5mm,
    minimum height=.48cm,align=center,line width=.5pt},
  statebox/.style={draw=steel,fill=white,rounded corners=.35mm,
    minimum height=.56cm,align=center,line width=.5pt,font=\figfont\scriptsize},
  flow/.style={->,draw=ink,line width=.75pt},
  condition/.style={->,draw=plum,line width=.8pt},
  supervise/.style={->,draw=teal,dashed,line width=.65pt},
  mutednote/.style={font=\figfont\footnotesize,text=subink,align=center}
]

\node[panel,minimum width=10.45cm,minimum height=3.85cm,anchor=south west] (pa) at (0,4.75) {};
\node[title] at (0.24,8.28) {(a) Learned operator splitting};

\node[font=\figfont\normalsize,anchor=west] at (0.45,7.72) {Geometry};
\airfoil{(0.40,7.34)}
\draw[plum,line width=.8pt] (2.20,7.40) -- (8.30,7.40);
\node[text=plum,font=\figfont\small] at (5.05,7.64) {conditioning};

\node[inner sep=0pt,draw=linegray,fill=white,minimum width=1.23cm,
  minimum height=1.12cm] (vin) at (0.92,6.16) {\velocityimage};
\node[font=\figfont\bfseries] at (0.92,6.91) {$u^n$};
\node[mutednote] at (0.92,5.10) {Velocity \\ Frame $n$};

\node[draw=plum,dashed,rounded corners=1.8mm,fill=plumlight!35,
  minimum width=7.15cm,minimum height=1.55cm] (gops) at (5.48,6.14) {};
\node[stage] (adv)  at (2.85,6.12) {Advection};
\node[stage] (diff) at (5.28,6.12) {Diffusion};
\node[stage] (proj) at (7.72,6.12) {Projection};
\node[circle,draw=ink,fill=plumlight,minimum size=.24cm,inner sep=0pt] (s1) at (4.02,6.12) {};
\node[circle,draw=ink,fill=plumlight,minimum size=.24cm,inner sep=0pt] (s2) at (6.46,6.12) {};
\node[circle,draw=ink,fill=plumlight,minimum size=.24cm,inner sep=0pt] (s3) at (8.88,6.12) {};
\node[font=\figfont\bfseries] at (4.02,6.52) {$u_{*}$};
\node[font=\figfont\bfseries] at (6.46,6.52) {$u_{**}$};
\node[font=\figfont\bfseries] at (9.78,6.47) {$u^{(n+1)}$};
\node[mutednote] at (9.65,5.82) {Frame $n{+}1$};

\draw[flow] (vin.east) -- (adv.west);
\draw[flow] (adv.east) -- (s1.west);
\draw[flow] (s1.east) -- (diff.west);
\draw[flow] (diff.east) -- (s2.west);
\draw[flow] (s2.east) -- (proj.west);
\draw[flow] (proj.east) -- (s3.west);
\draw[flow] (s3.east) -- ++(.70,0);
\foreach \x in {2.85,5.28,7.72}{\draw[condition] (\x,7.40) -- (\x,6.52);}
\foreach \x in {3.53,5.96,8.40}{\node[fill=plum,text=white,font=\figfont\bfseries\tiny,inner sep=1.2pt] at (\x,6.40) {[G]};}

\node[panel,minimum width=10.45cm,minimum height=4.30cm,anchor=south west] (pb) at (0,0.25) {};
\node[title] at (0.24,4.18) {(b) Training(MC Label)};

\foreach \id/\x/\slot/\label in {
  A/4.02/\mcadvectionimage/MC advection,
  B/6.46/\mcdiffusionimage/MC diffusion,
  C/8.88/\mcprojectionimage/MC projection}{
    \node[circle,draw=teal,fill=teallight,minimum size=.48cm,inner sep=0pt,
      line width=.50pt,font=\figfont\scriptsize] (loss\id) at (\x,3.45) {$\mathcal L$};
    \node[inner sep=0pt,draw=linegray,fill=white,line width=.35pt,
      minimum width=1.35cm,minimum height=1.35cm] (img\id) at (\x,2.15) {\slot};
    \node[mutednote,anchor=east] at (\x-.84,2.15) {MC\\target};
    \node[mcbox] (mc\id) at (\x,0.95) {\label};
    \draw[flow] (mc\id.north) -- (img\id.south);
    \draw[supervise] (img\id.north) -- (loss\id.south);
  }
\draw[supervise] (s1.south) -- (lossA.north);
\draw[supervise] (s2.south) -- (lossB.north);
\draw[supervise] (s3.south) -- (lossC.north);

\node[draw=linegray,fill=white,rounded corners=.7mm,anchor=south west,
  minimum width=2.48cm,minimum height=.68cm,align=left,font=\figfont\tiny]
  at (0.28,0.46) {\tikz\draw[flow] (0,0)--(.65,0);\quad forward\\[-1pt]
                     \tikz\draw[supervise] (0,0)--(.65,0);\quad supervision};

\node[panel,minimum width=8.30cm,minimum height=8.35cm,anchor=south west] (pc) at (10.70,0.25) {};
\node[font=\figfont\bfseries\small,anchor=west] at (10.98,8.30)
  {(c) Inside a geometry-conditioned operator \textcolor{plum}{[G]}};

\node[draw=plum,dashed,rounded corners=1.4mm,fill=plumlight!32,
  minimum width=7.60cm,minimum height=5.02cm,anchor=north] (gb) at (14.85,7.92) {};
\node[anchor=west,font=\figfont\bfseries\small,text=plum]
  at (11.18,7.63) {Geometry branch};

\node[softbox,minimum width=2.42cm] (sdf) at (12.75,7.02) {};
\node[font=\figfont\small,anchor=west] at ($(sdf.west)+(0.22,0)$) {SDF};
\begin{scope}[shift={(12.92,6.96)},scale=.54]
  \airfoil{(0,0)}
\end{scope}
\node[softbox,minimum width=2.68cm] (coords) at (16.70,7.02) {Coordinates $(x,y)$};
\node[softbox,draw=sand,fill=sandlight,minimum width=3.25cm] (liftg) at (14.73,5.91)
  {Lift + grid embedding};

\coordinate (inputleft)  at (12.75,6.43);
\coordinate (inputright) at (16.70,6.43);
\coordinate (inputjoin)  at (14.73,6.43);
\draw[draw=ink,line width=.70pt] (sdf.south) -- (inputleft);
\draw[draw=ink,line width=.70pt] (coords.south) -- (inputright);
\draw[draw=ink,line width=.70pt] (inputleft) -- (inputright);
\draw[flow] (inputjoin) -- (liftg.north);

\node[draw=plum,fill=plumlight,rounded corners=.65mm,
  minimum width=5.25cm,minimum height=1.10cm] (genc) at (14.73,4.75) {};
\node[font=\figfont\bfseries\small,text=plum] at (14.73,5.06) {Geometry encoder};
\node[draw=sand,fill=sandlight,minimum width=4.12cm,minimum height=.52cm,
  font=\figfont\scriptsize] at (14.50,4.55) {Spectral conv + $1{\times}1$ skip};
\node[circle,fill=plum,text=white,minimum size=.45cm,inner sep=0pt,
  font=\figfont\scriptsize] at (16.91,4.55) {$\times4$};
\draw[flow] (liftg.south) -- (genc.north);

\node[softbox,minimum width=2.75cm] (mlp) at (14.73,3.42) {Pointwise MLP};

\draw[condition] (genc.south) -- (mlp.north);
\node[text=plum,font=\figfont\small\bfseries,anchor=west]
  at (15.02,3.95) {Multiscale features};

\node[draw=steel,dashed,rounded corners=1.4mm,fill=steellight!42,
  minimum width=7.60cm,minimum height=2.02cm,anchor=south] (sb) at (14.85,0.50) {};
\node[anchor=west,font=\figfont\bfseries\small,text=steel]
  at (11.18,2.28) {State operator};

\node[statebox,minimum width=.90cm] (vel) at (11.85,1.55) {Velocity};
\node[statebox,minimum width=.86cm] (pad) at (13.20,1.55) {Lift\\+ pad};
\node[statebox,fill=steellight,minimum width=1.72cm] (spec) at (14.73,1.55)
  {Spectral conv\\+ $1{\times}1$ skip};
\node[circle,fill=steel,text=white,minimum size=.42cm,inner sep=0pt,
  font=\figfont\tiny] at (15.31,1.25) {$\times4$};
\node[statebox,minimum width=1.02cm] (map) at (16.55,1.55) {Output\\mapping};
\node[circle,draw=ink,fill=white,minimum size=.36cm,inner sep=0pt] (plus) at (17.45,1.55) {$+$};
\node[statebox,minimum width=.82cm] (out) at (18.30,1.55) {Output};

\draw[flow] (vel.east) -- (pad.west);
\draw[flow] (pad.east) -- (spec.west);
\draw[flow] (spec.east) -- (map.west);
\draw[flow] (map.east) -- (plus.west);
\draw[flow] (plus.east) -- (out.west);

\draw[condition] (mlp.south) -- (spec.north);
\node[text=plum,font=\figfont\small\bfseries,anchor=west]
  at (15.02,2.70) {Gates};

\draw[draw=ink,line width=.55pt] (vel.south) -- ++(0,-.40) -| (plus.south);
\node[font=\figfont,text=subink,fill=steellight!42,inner sep=.7pt]
  at (15.15,.65) {residual bypass};

\end{tikzpicture}

%% file: iclr2027_conference.bib
@article{LAN2023112343,
title = {DOSnet as a non-black-box PDE solver: When deep learning meets operator splitting},
journal = {Journal of Computational Physics},
volume = {491},
pages = {112343},
year = {2023},
issn = {0021-9991},
doi = {https://doi.org/10.1016/j.jcp.2023.112343},
url = {https://www.sciencedirect.com/science/article/pii/S0021999123004382},
author = {Yuan Lan and Zhen Li and Jie Sun and Yang Xiang}
}

@InProceedings{pmlr-v300-gopakumar26a,
  title = 	 { Learning Physical Operators using Neural Operators },
  author =       {Gopakumar, Vignesh and Gray, Ander and Giles, Daniel and Zanisi, Lorenzo and Kusner, Matt J. and Betcke, Timo and Pamela, Stanislas and Deisenroth, Marc Peter},
  booktitle = 	 {Proceedings of The 29th International Conference on Artificial Intelligence and Statistics},
  pages = 	 {3223--3231},
  year = 	 {2026},
  editor = 	 {Khan, Emtiyaz and Li, Yingzhen and Solin, Arno and Ramdas, Aaditya},
  volume = 	 {300},
  series = 	 {Proceedings of Machine Learning Research},
  month = 	 {02--05 May},
  publisher =    {PMLR},
  url = 	 {https://proceedings.mlr.press/v300/gopakumar26a.html}
}

@inproceedings{serrano2026disco,
  title     = {Test-time Generalization for Physics through Neural Operator Splitting},
  author    = {Serrano, Louis and Han, Jiequn and Oyallon, Edouard and Ho, Shirley and Morel, Rudy},
  booktitle = {International Conference on Machine Learning (ICML)},
  year      = {2026}
}

@inproceedings{
sun2026origo,
title={Origo: Interpretable Multi-physics {PDE} Foundation Model through Neural Operator Splitting},
author={Li Sun and Hongbo Lv and Zhikai Jiang and Zhongtian Sun and Lanxu Yang and Philip S. Yu},
booktitle={Forty-third International Conference on Machine Learning},
year={2026},
url={https://openreview.net/forum?id=nBDuY2PjTY}
}

@inproceedings{hmida2026compositional,
  title     = {Compositional Neural Operators for Multi-Dimensional Fluid Dynamics},
  author    = {Hamda Hmida and Hsiu-Wen Chang and Youssef Mesri},
  booktitle = {International Conference on Learning Representations},
  year      = {2026}
}

@InProceedings{pmlr-v202-helwig23a,
  title = 	 {Group Equivariant {F}ourier Neural Operators for Partial Differential Equations},
  author =       {Helwig, Jacob and Zhang, Xuan and Fu, Cong and Kurtin, Jerry and Wojtowytsch, Stephan and Ji, Shuiwang},
  booktitle = 	 {Proceedings of the 40th International Conference on Machine Learning},
  pages = 	 {12907--12930},
  year = 	 {2023},
  editor = 	 {Krause, Andreas and Brunskill, Emma and Cho, Kyunghyun and Engelhardt, Barbara and Sabato, Sivan and Scarlett, Jonathan},
  volume = 	 {202},
  series = 	 {Proceedings of Machine Learning Research},
  month = 	 {23--29 Jul},
  publisher =    {PMLR},
  url = 	 {https://proceedings.mlr.press/v202/helwig23a.html}
}

@inproceedings{
puny2022frame,
title={Frame Averaging for Invariant and Equivariant Network Design},
author={Omri Puny and Matan Atzmon and Edward J. Smith and Ishan Misra and Aditya Grover and Heli Ben-Hamu and Yaron Lipman},
booktitle={International Conference on Learning Representations},
year={2022},
url={https://openreview.net/forum?id=zIUyj55nXR}
}

@inproceedings{NEURIPS2023_3b5c7c9c,
 author = {Kim, Jinwoo and Nguyen, Dat and Suleymanzade, Ayhan and An, Hyeokjun and Hong, Seunghoon},
 booktitle = {Advances in Neural Information Processing Systems},
 doi = {10.52202/075280-0816},
 editor = {A. Oh and T. Naumann and A. Globerson and K. Saenko and M. Hardt and S. Levine},
 pages = {18582--18612},
 publisher = {Curran Associates, Inc.},
 title = {Learning Probabilistic Symmetrization for Architecture Agnostic Equivariance},
 url = {https://proceedings.neurips.cc/paper_files/paper/2023/file/3b5c7c9c5c7bd77eb73d0baec7a07165-Paper-Conference.pdf},
 volume = {36},
 year = {2023}
}

@article{DBLP:journals/corr/abs-2511-09573,
  author       = {Valentino F. Foit and
                  David W. Hogg and
                  Soledad Villar},
  title        = {Group Averaging for Physics Applications: Accuracy Improvements at
                  Zero Training Cost},
  journal      = {CoRR},
  volume       = {abs/2511.09573},
  year         = {2025},
  url          = {https://doi.org/10.48550/arXiv.2511.09573},
  doi          = {10.48550/ARXIV.2511.09573},
  eprinttype   = {arXiv},
  eprint       = {2511.09573},
  bibsource    = {dblp computer science bibliography, https://dblp.org}
}

@inproceedings{DBLP:conf/iclr/BrandstetterWW22,
  author       = {Johannes Brandstetter and
                  Daniel E. Worrall and
                  Max Welling},
  title        = {Message Passing Neural {PDE} Solvers},
  booktitle    = {The Tenth International Conference on Learning Representations, {ICLR}
                  2022, Virtual Event, April 25-29, 2022},
  publisher    = {OpenReview.net},
  year         = {2022},
  url          = {https://openreview.net/forum?id=vSix3HPYKSU},
  bibsource    = {dblp computer science bibliography, https://dblp.org}
}

@article{JMLR:v24:23-0064,
  author  = {Zongyi Li and Daniel Zhengyu Huang and Burigede Liu and Anima Anandkumar},
  title   = {Fourier Neural Operator with Learned Deformations for PDEs on General Geometries},
  journal = {Journal of Machine Learning Research},
  year    = {2023},
  volume  = {24},
  number  = {388},
  pages   = {1--26},
  url     = {http://jmlr.org/papers/v24/23-0064.html}
}

@inproceedings{NEURIPS2023_940a7634,
 author = {Liu, Ning and Jafarzadeh, Siavash and Yu, Yue},
 booktitle = {Advances in Neural Information Processing Systems},
 doi = {10.52202/075280-2053},
 editor = {A. Oh and T. Naumann and A. Globerson and K. Saenko and M. Hardt and S. Levine},
 pages = {47438--47450},
 publisher = {Curran Associates, Inc.},
 title = {Domain Agnostic Fourier Neural Operators},
 url = {https://proceedings.neurips.cc/paper_files/paper/2023/file/940a7634dab556b67af15bacd337f7db-Paper-Conference.pdf},
 volume = {36},
 year = {2023}
}

@inproceedings{NEURIPS2023_70518ea4,
 author = {Li, Zongyi and Kovachki, Nikola and Choy, Chris and Li, Boyi and Kossaifi, Jean and Otta, Shourya and Nabian, Mohammad Amin and Stadler, Maximilian and Hundt, Christian and Azizzadenesheli, Kamyar and Anandkumar, Animashree},
 booktitle = {Advances in Neural Information Processing Systems},
 doi = {10.52202/075280-1556},
 editor = {A. Oh and T. Naumann and A. Globerson and K. Saenko and M. Hardt and S. Levine},
 pages = {35836--35854},
 publisher = {Curran Associates, Inc.},
 title = {Geometry-Informed Neural Operator for Large-Scale 3D PDEs},
 url = {https://proceedings.neurips.cc/paper_files/paper/2023/file/70518ea42831f02afc3a2828993935ad-Paper-Conference.pdf},
 volume = {36},
 year = {2023}
}

@article{ZENG2025118022,
title = {Point cloud neural operator for parametric PDEs on complex and variable geometries},
journal = {Computer Methods in Applied Mechanics and Engineering},
volume = {443},
pages = {118022},
year = {2025},
issn = {0045-7825},
doi = {https://doi.org/10.1016/j.cma.2025.118022},
url = {https://www.sciencedirect.com/science/article/pii/S0045782525002944},
author = {Chenyu Zeng and Yanshu Zhang and Jiayi Zhou and Yuhan Wang and Zilin Wang and Yuhao Liu and Lei Wu and Daniel Zhengyu Huang}
}

@article{56b77649b60443079266e65254c18cd0,
title = "Variationally mimetic operator networks",
author = "Dhruv Patel and Deep Ray and Abdelmalik, \{Michael R.A.\} and Hughes, \{Thomas J.R.\} and Oberai, \{Assad A.\}",
year = "2024",
month = feb,
day = "1",
doi = "10.1016/j.cma.2023.116536",
language = "English",
volume = "419",
journal = "Computer Methods in Applied Mechanics and Engineering",
issn = "0045-7825",
publisher = "Elsevier B.V.",
}

@article{cf62894765ac45ccbdc080450ad4aef3,
title = "Neural Green{\textquoteright}s operators for parametric partial differential equations",
author = "H.A. Melchers and J.H.M. Prins and M.R.A. Abdelmalik",
year = "2026",
month = jun,
day = "15",
doi = "10.1016/j.cma.2026.118893",
language = "English",
volume = "455",
journal = "Computer Methods in Applied Mechanics and Engineering",
issn = "0045-7825",
publisher = "Elsevier B.V.",
}

@inproceedings{NEURIPS2025_448abd48,
 author = {Yoo, Seungwoo and Yeo, Kyeongmin and Hwang, Jisung and Sung, Minhyuk},
 booktitle = {Advances in Neural Information Processing Systems},
 doi = {10.52202/085713-1599},
 editor = {D. Belgrave and C. Zhang and H. Lin and R. Pascanu and P. Koniusz and M. Ghassemi and N. Chen},
 pages = {47961--47982},
 publisher = {Curran Associates, Inc.},
 title = {Neural Green’s Functions},
 url = {https://proceedings.neurips.cc/paper_files/paper/2025/file/448abd486677165ceedfa790e9a61802-Paper-Conference.pdf},
 volume = {38, Main Conference},
 year = {2025}
}

@inproceedings{alesiani2022hyperfno,
 author = {Alesiani, Francesco and Takamoto, Makoto and Niepert, Mathias},
 booktitle = {NeurIPS 2022 Workshop on Machine Learning and Physical Sciences},
 title = {HyperFNO: Improving the generalization behavior of Fourier Neural Operators},
 year = {2022}
}

@inproceedings{ICLR2026_c253f6fe,
 author = {Jing, Cheng and Balasuriya Mudiyanselage, Uvini and Verma, Abhishek and Bera, Kallol and Rauf, Shahid and Lee, Kookjin},
 booktitle = {International Conference on Learning Representations},
 editor = {C. Vondrick and B. Hariharan and C. Raffel and L. Pinto and D. Yang and A. Faust},
 pages = {119621--119651},
 title = {Extending Fourier Neural Operators for Modeling Parameterized and Coupled PDEs},
 url = {https://proceedings.iclr.cc/paper_files/paper/2026/file/c253f6fe5cc2bd1905427da63ac37917-Paper-Conference.pdf},
 volume = {2026},
 year = {2026}
}

@article{weller1998tensorial,
  title={A tensorial approach to computational continuum mechanics using object-oriented techniques},
  author={Weller, Henry G and Tabor, Gavin and Jasak, Hrvoje and Fureby, Christer},
  journal={Computers in physics},
  volume={12},
  number={6},
  pages={620--631},
  year={1998},
  publisher={American Institute of Physics}
}

@manual{openfoamSolutionControl,
  author = {{CFD Direct}},
  title = {{OpenFOAM} v13 User Guide: Solution and Algorithm Control},
  year = {2025},
  url = {https://doc.cfd.direct/openfoam/user-guide-v13/fvsolution},
  note = {Section 4.6, accessed September 25, 2026}
}

@inproceedings{Sugimoto:2024:VelMCFluids,
author = {Sugimoto, Ryusuke and Batty, Christopher and Hachisuka, Toshiya},
title = {Velocity-Based Monte Carlo Fluids},
year = {2024},
isbn = {9798400705250},
publisher = {Association for Computing Machinery},
address = {New York, NY, USA},
url = {https://doi.org/10.1145/3641519.3657405},
doi = {10.1145/3641519.3657405},
booktitle = {ACM SIGGRAPH 2024 Conference Papers},
articleno = {8},
numpages = {11},
location = {Denver, CO, USA},
series = {SIGGRAPH '24}
}

@inproceedings{
li2021fourier,
title={Fourier Neural Operator for Parametric Partial Differential Equations},
author={Zongyi Li and Nikola Borislavov Kovachki and Kamyar Azizzadenesheli and Burigede liu and Kaushik Bhattacharya and Andrew Stuart and Anima Anandkumar},
booktitle={International Conference on Learning Representations},
year={2021},
url={https://openreview.net/forum?id=c8P9NQVtmnO}
}

@article{chorin1968numerical,
  title   = {Numerical solution of the {Navier--Stokes} equations},
  author  = {Chorin, Alexandre Joel},
  journal = {Mathematics of Computation},
  volume  = {22}, number = {104}, pages = {745--762}, year = {1968},
  doi     = {10.1090/S0025-5718-1968-0242392-2}
}

@InProceedings{Koch_2019_CVPR,
author = {Koch, Sebastian and Matveev, Albert and Jiang, Zhongshi and Williams, Francis and Artemov, Alexey and Burnaev, Evgeny and Alexa, Marc and Zorin, Denis and Panozzo, Daniele},
title = {ABC: A Big CAD Model Dataset for Geometric Deep Learning},
booktitle = {Proceedings of the IEEE/CVF Conference on Computer Vision and Pattern Recognition (CVPR)},
month = {June},
year = {2019}
}

@incollection{macnamara2017operator,
  title={Operator splitting},
  author={MacNamara, Shev and Strang, Gilbert},
  booktitle={Splitting methods in communication, imaging, science, and engineering},
  pages={95--114},
  year={2017},
  publisher={Springer}
}

@article{POPINET20095838,
title = {An accurate adaptive solver for surface-tension-driven interfacial flows},
journal = {Journal of Computational Physics},
volume = {228},
number = {16},
pages = {5838-5866},
year = {2009},
issn = {0021-9991},
doi = {https://doi.org/10.1016/j.jcp.2009.04.042},
url = {https://www.sciencedirect.com/science/article/pii/S002199910900240X},
author = {Stéphane Popinet}
}

@article{Orszag_Kells_1980, title={Transition to turbulence in plane Poiseuille and plane Couette flow}, volume={96}, DOI={10.1017/S0022112080002066}, number={1}, journal={Journal of Fluid Mechanics}, author={Orszag, Steven A. and Kells, Lawrence C.}, year={1980}, pages={159–205}}
